\documentclass[a4paper,fleqn]{cas-dc}
\usepackage{algorithm}
\usepackage{algpseudocode}
\usepackage{hyperref}
\usepackage{natbib}
\setcitestyle{style=authoryear, maxcitenames=1}

\begin{document}
\let\printorcid\relax
\let\WriteBookmarks\relax
\def\floatpagepagefraction{1}
\def\textpagefraction{.001}

\shorttitle{A Benchmark Dataset and CMA model}

\shortauthors{Shibo Zhou et~al.}

\title [mode = title]{Enhancing SNN-based Spatio-Temporal Learning: A Benchmark Dataset and Cross-Modality Attention Model}

\tnotetext[0]{This work was supported by the National Natural Science Foundation of China under Grant 62236007 and 62276235.}

\tnotetext[0]{The source code and DVS-SLR dataset are available at \href{https://github.com/JasonKitty/DVS-SLR}{our GitHub Repository}.}



\author[1]{Shibo Zhou}
\ead{shibo.zhou@zhejianglab.com}
\author[2]{Bo Yang}
\ead{yangboak@icloud.com}

\author[3]{Mengwen Yuan}
\ead{yuanmw@zhejianglab.com}

\author[2]{Runhao Jiang}
\ead{rhjiang@zju.edu.cn}

\author[4]{Rui Yan}
\ead{ryan@zjut.edu.cn}

\author[2,5]{Gang Pan}
\ead{gpan@zju.edu.cn}

\author[2,5]{Huajin Tang}
\cormark[1]
\ead{htang@zju.edu.cn}

\affiliation[1]{organization={Research Center for Data Hub and Security, Zhejiang Lab},    city={Hangzhou},
    country={China}}
    
\affiliation[2]{organization={College of Computer Science and Technology, Zhejiang University},    city={Hangzhou},
    country={China}}
    
\affiliation[3]{organization={Research Center for High Efficiency Computing System, Zhejiang Lab},    city={Hangzhou},
    country={China}}
    
\affiliation[4]{organization={College of Computer Science and Technology, Zhejiang University of Technology},
    city={Hangzhou},
    country={China}}
    
\affiliation[5]{organization={The State Key Lab of Brain-Machine Intelligence, Zhejiang University},    city={Hangzhou},
    country={China}}













\cortext[cor1]{Corresponding author}



\begin{abstract}
Spiking Neural Networks (SNNs), renowned for their low power consumption, brain-inspired architecture, and spatio-temporal representation capabilities, have garnered considerable attention in recent years. Similar to Artificial Neural Networks (ANNs), high-quality benchmark datasets are of great importance to the advances of SNNs. However, our analysis indicates that many prevalent neuromorphic datasets lack strong temporal correlation, preventing SNNs from fully exploiting their spatio-temporal representation capabilities. Meanwhile, the integration of event and frame modalities offers more comprehensive visual spatio-temporal information. Yet, the SNN-based cross-modality fusion remains underexplored.

In this work, we present a neuromorphic dataset called DVS-SLR that can better exploit the inherent spatio-temporal properties of SNNs. Compared to existing datasets, it offers advantages in terms of higher temporal correlation, larger scale, and more varied scenarios. In addition, our neuromorphic dataset contains corresponding frame data, which can be used for developing SNN-based fusion methods. By virtue of the dual-modal feature of the dataset, we propose a Cross-Modality Attention (CMA) based fusion method. The CMA model efficiently utilizes the unique advantages of each modality, allowing for SNNs to learn both temporal and spatial attention scores from the spatio-temporal features of event and frame modalities, subsequently allocating these scores across modalities to enhance their synergy. Experimental results demonstrate that our method not only improves recognition accuracy but also ensures robustness across diverse scenarios.
\end{abstract}




\maketitle

\section{Introduction}
The spiking neural network (SNN), as a promising brain-inspired computational model, has garnered considerable attention in recent years. Event cameras, a type of neuromorphic vision sensor, generate a stream of asynchronous events only after detecting intensity changes. This characteristic aligns naturally with the event-driven, spike-based, and spatio-temporal representation property of SNNs. 

The datasets captured by event cameras are known as neuromorphic vision datasets. A suitable benchmark dataset is crucial for fairly evaluating improved algorithms and advancing the development of SNNs. Nonetheless, many studies have highlighted the shortcomings of current SNN datasets. 
\citep{RN30} compared the performance of SNNs and RNNs under different workloads and found that SNNs have a distinct advantage in high temporal resolution workloads—a feature not supported by many existing datasets. 
\citep{RN29} conducted unsupervised RD-STDP (Reward-Dependent Spike-Timing-Dependent Plasticity) and STDP-tempotron methods on the N-MNIST dataset and discovered that temporal information does not play a significant role in classification for N-MNIST; even when the event stream is accumulated into a single frame, the accuracy does not decrease significantly.
\citep{RN31} found that the performance of SNNs on some neuromorphic datasets was even inferior to ANNs and underscored the urgent need to create more SNN-oriented datasets.

In this work, we designed two experiments called Spike Timing Confusion and Temporal Information Elimination to quantify and analyze the importance of temporal information. Our analysis indicates that many prevalent neuromorphic datasets lack sufficient temporal correlation, preventing SNNs from fully exploiting their spatio-temporal representation capabilities. Building on these findings, we believe that an appropriate benchmark dataset would enable SNNs to fully exploit the characteristics of spatio-temporal representation and rapid response, allowing SNNs to demonstrate their distinctiveness \citep{li2022n}. 
To this end, we design and record a sign language action recognition dataset using the DAVIS346 event camera. To date, it is the most extensive action recognition neuromorphic dataset, featuring 21 sign language action segments performed by 43 participants under three lighting conditions and two position conditions, with a total recording time exceeding 9 hours. To validate its optimal use of spatio-temporal properties of SNNs, we carry out Spike Timing Confusion and Temporal Information Elimination experiments. The former jumbles the spike timings of the input event stream, feeding the model erroneous temporal information, while the latter accumulates the entire event stream spatially into one frame, preventing the model from accessing temporal information. Experimental results suggest our dataset has higher temporal correlation and increased recognition difficulty. We underscore that it is not necessary for researchers to pursue the marginal performance improvement of SNN on ANN-oriented datasets, but to tackle some problems that render more challenges to ANNs. Our dataset can better exploit the inherent properties of SNNs, thus paving the way for fair comparisons between SNN algorithms and exploration of spatio-temporal properties. 

On the other hand, event modality is difficult to represent background and texture information. Integrating event modality with traditional color frame modality can facilitate a more holistic understanding of a visual scene. Such integration has been successfully applied to target tracking \citep{luo2022conversion, wang2021visevent}, image reconstruction \citep{paikin2021efi, tulyakov2022time}, SLAM \citep{weikersdorfer2014event, SLAM}, and more. However, most of these applications are based on ANNs and ignore the rich temporal information captured in the event modality. Due to the lack of dual-modal datasets and the inherent complexities of the neurodynamics, the development of SNN-based modality fusion techniques has been impeded.

As such, we synchronously recorded the corresponding color frame data when capturing the neuromorphic dataset. Based on this dual-modal dataset, we introduce a Cross-Modality Attention (CMA) module and establish a baseline for SNN-based fusion. This module is designed end-to-end and can be seamlessly integrated into SNN framework. It comprises two distinct components: the spatial-wise CMA and the temporal-wise CMA. For the spatial-wise CMA, it first computes the spatial-wise spike rate of the frame features. The spike rate can be interpreted as the degree of response to specific information. Subsequently, a learnable 2D nonlinear convolutional mapping is executed, producing spatial attention scores. These scores are then cross-fused with event features, resulting in spatially enhanced features. In parallel, the temporal-wise CMA undergoes a similar procedure. It begins by determining the temporal-wise spike rate of the event features. Subsequently, a learnable nonlinear 1D fully-connected mapping is employed to yield temporal attention scores, which are then cross-fused with frame features to craft temporally enhanced features. We conducted detailed experiments under various latency parameters and scenario settings. Our experimental results show that the integration of CMA not only improves the recognition accuracy but also enables the model maintaining robustness under diverse conditions. Furthermore, The efficacy of the CMA module is confirmed through a comprehensive ablation study.

In summary, our contributions are as follows:
\begin{enumerate}[(1)]
\item 
We introduce a benchmark neuromorphic dataset. Compared to other commonly used datasets, it can better enable SNNs to exploit their inherent properties, especially in terms of spatio-temporal representation. Additionally, it encompasses frame data synchronized with the event stream, making it suitable for developing SNN-based fusion techniques. 

\item
We delve deeply into the inadequacies of current neuromorphic datasets in terms of temporal correlation, an aspect that many previous studies have highlighted but not thoroughly investigated. Through conducting Spike Timing Confusion and Temporal Information Elimination experiments, we substantiate that our dataset exhibits a higher degree of temporal correlation, establishing it as a appropriate benchmark for SNN evaluations.

\item
We introduce a Cross-Modality Attention (CMA) model that captures both temporal and spatial attention scores from the spatio-temporal features in event and frame modalities. These scores are then cross-allocated to the other modality, enhancing the synergy between them.
\end{enumerate}

\section{Related Works}
\subsection{Spiking Neural Network}

Unlike ANNs, each neuron in an SNN accumulates incoming signals over time. When the accumulated signal surpasses a given threshold, the neuron 'fires', or produces a binary spike, and resets. This dynamic allows SNNs to encode and process information both spatially and temporally \citep{gerstner2014neuronal}. Moreover, the inherent event-driven characteristics of SNNs allow for faster response and lower power consumption, making them highly suitable for tasks requiring real-time responses \citep{bouvier2019spiking}.

However, the introduction of spiking neurons results in the non-differentiable nature of SNNs, rendering conventional backpropagation optimization unsuitable for training SNNs directly. To address this issue, researchers have explored two main approaches. 

One approach involves initially training the desired model using ANNs. Given a pre-trained ANN, ANN-SNN conversion involves substituting the activation function with spiking neurons, accompanied by meticulous parameter adaptation, to replicate the continuous feature representation via spiking neurons \citep{diehl2015fast, diehl2016conversion, han2023symmetric, qu2023spiking}. Some studies optimize the intrinsic spatiotemporal dynamics of spiking neurons to achieve low latency and sparse computing \citep{zhang2023low, zhang2021event}. A large amount of recent work has demonstrated that the performance of converted SNNs is comparable to ANNs over large-scale datasets \citep{NEURIPS2021_afe43465, bu2023optimal, hao2023reducing, RN79, wang2023toward}. However, the converted SNNs are unable to process continuous inputs due to the limitations of the original ANN input formats.

Another approach aims to train SNNs directly by introducing the concept of surrogate gradients, which make the non-differentiable process of spike firing derivable. A significant advancement in this domain is the Spatio-Temporal Back Propagation (STBP) method \citep{wu2019direct}, which combines the principles of Back Propagation Through Time \citep{werbos1990backpropagation} with surrogate gradient techniques, showing promising results. Further developments in neuron models \citep{fang2021incorporating, yu2022improving, ma2023Exploiting}, network architectures \citep{hu2021spiking, fang2021deep, yao2021temporal, feng2022multi}, and training techniques \citep{xie2023event, hu2021spiking, zheng2021going, yang2022training, wang2023adaptive, gu2019stca} underscore the immense potential of SNNs. This direct training approach is inherently adept at managing continuous inputs, potentially paving the way for advancements in fields such as action recognition \citep{liu2021event, gao2023action}, target tracking \citep{cao2015spiking}, mapless navigation \citep{yang2023spiking}, trajectory prediction \citep{debat2021event}, object detection \citep{yu2024spikingvit, yuan2024trainable},and more.

\subsection{Neuromorphic Datasets}
Neuromorphic datasets are specialized datasets developed to train and test neuromorphic computing systems, such as SNNs. These datasets encode information in an event-driven manner, where each "event" signifies a spike generated by a neuron at a specific moment in time. These datasets can encompass various types of data including but not limited to images, sounds \citep{anumula2018feature}, or other sensor data \citep{see2020st, vanarse2022application}. This discussion is limited to neuromorphic vision datasets.

Neuromorphic vision datasets can generally be divided into three categories based on their source and generation methods: (i) Simulated Datasets, (ii) Converted Datasets, and (iii) Real-World Datasets. 

Among them, the simulated datasets are directly generated from existing video streams (or images) using specific difference algorithms \citep{rebecq2018esim} or frame-based generation algorithms \citep{hu2021v2e} to create a neuromorphic version of the original data. These datasets are generally used to provide additional dynamic information to assist in tasks such as tracking \citep{wang2021visevent, zhang2023frame}, depth estimation \citep{RN36}, optical flow estimation \citep{li2023blinkflow}, image reconstruction \citep{wang2019event}, and more. However, a significant downside is the inability to accurately emulate event-driven characteristics, attributed to the limited frame rate of conventional videos. Consequently, it fails to correctly reflect real-time spatio-temporal information, thereby offering restricted support in fostering the progress of SNNs.

The converted datasets refer to neuromorphic datasets formulated from well-established datasets in the ANN domain; notable examples include N-MNIST \citep{RN32} derived from MNIST, N-ImageNet \citep{RN42} from ImageNet, and DVS-CIFAR10 \citep{RN35} from CIFAR10, among others \citep{RN43, hu2016dvs}.  Generally, the creation of these datasets entails initially stabilizing the original images, followed by moving the event camera along a predetermined path, thereby generating a corresponding event stream version. However, these datasets, despite meeting the formal prerequisites for SNNs, are not genuinely neuromorphic in nature. Our research (see Section \ref{Dataset_Analysis}) indicates that the precise spike timing of these datasets bears negligible influence on the accuracy, and the total elimination of temporal information does not cause a marked decrease in performance. Hence, these datasets fall short in fully harnessing the spatio-temporal representation potential inherent in SNNs, making them unsuitable as benchmark datasets for SNN evaluations. This standpoint is echoed by numerous other studies in the field \citep{RN29, RN30, RN31}.

The real-world datasets are recorded by event cameras in actual scenes. Compared to the aforementioned methods, the data acquired in this manner is authentic and natural, directly reflecting the motion of objects and the spatio-temporal characteristics in the real environments, offering immense value for the advancement of SNNs. This type of datasets has already been widely used for gesture recognition \citep{amir2017low, RN68, RN39, liu2021event, dong2022event}, and object classification \citep{serrano2015poker, sironi2018hats}. However, with the rapid advancements in SNNs, the existing datasets have become inadequate for fostering the further development of SNNs. For instance, the most frequently utilized DVS-Gesture dataset \citep{amir2017low} is now too simplistic for the current approaches of SNNs, allowing an easy attainment of about 95\% recognition accuracy. Moreover, these datasets are unimodal, offering no assistance in the research of spatio-temporal fusion techniques in SNNs and lacking a corresponding color frame based baseline for comparison.

\subsection{Collaboration of Event and Frame Modalities}
Traditional frame-based cameras excel at providing rich details concerning background and texture. However, they suffer from motion blur when capturing high-speed moving objects and have a lower dynamic range. Event cameras can overcome these shortcomings by offering a higher temporal resolution and dynamic range, supplying complementary dynamic information to the frame stream \citep{gehrig2021combining}.

In the early stages, researchers envisioned the potential of fusing these two modalities for video processing \citep{leow2015machine}. Following this, with the advancement of deep neural networks, researchers began to leverage the benefits of this fusion in more specialized applications such as SLAM \citep{weikersdorfer2014event, SLAM}, image reconstruction \citep{paikin2021efi, tulyakov2022time}, monocular depth estimation \citep{gehrig2021combining}, object tracking \citep{luo2022conversion, wang2021visevent}, object localization \citep{lele2022bio}, semantic segmentation \citep{natan2021semantic}, and image enhancement \citep{bi2023non}. 

Nevertheless, the majority of these cases utilize the event stream merely as an auxiliary modality, aiming to alleviate the deficiencies in the frame modality, particularly their low temporal resolution and limited dynamic range. These approaches predominantly rely on ANNs and often necessitate complex preprocessing and crafted designs, bypassing the direct engagement with spatio-temporal feature modeling. In this work, we aim to propose an end-to-end fusion strategy, which is tailored to the unique characteristics of each modality and facilitates seamless integration within the SNN framework.

\section{DVS-SLR Benchmark Dataset}
With the rapid development of neuron models, network architectures, and training techniques, there is an urgent need for SNN-oriented benchmark datasets \citep{RN31}. We align with \citep{RN29, RN30, li2022n} in the viewpoint that most of the so-called "neuromorphic" datasets are not quite "neuromorphic". Although their data format aligns with the event-driven nature of SNNs, the temporal information contained may not be significant enough. Moreover, these datasets are uni-modal, offering no assistance in the research of modality fusion techniques in SNNs and lacking a corresponding color-frame based baseline for comparison.

As such, we designed and recorded a dual-modal neuromorphic dataset. This dataset serves two purposes. When its event modality is used alone, it can act as a more suitable benchmark dataset for SNN evaluation, as it has higher temporal correlation and can fully exploit the spatio-temporal nature of SNNs. When used in its dual-modal form, it can assist in the development and validation of SNN-based fusion methods.

In this section, we will introduce the dataset's design, the recording process, and its analysis.

\subsection{Dataset Design} 
The output of a neuron in a SNN at any given timestep depends on the current input and the state of the neuron. The neuron's state carries information from the past. This computational paradigm endows SNNs with an intrinsic ability for temporal modeling. Additionally, information is transmitted in the form of spikes (0 or 1) in SNNs, enabling them to operate at lower power consumption and with faster response times. Considering these attributes, an ideal benchmark dataset or task should highlight the importance of both temporal and spatial information \citep{li2022n, RN29}, and the input data should be continuous. Additionally, the benchmark task should be straightforward and simple, without requiring complex data preprocessing.

Action recognition, reliant on subtle variations in spatio-temporal sequences and sparse data processing, naturally aligns with the properties of SNNs, making it an ideal benchmark task for SNNs. While there are existing neuromorphic datasets for action recognition \citep{amir2017low, RN68, RN39, liu2021event}, they fall short in scale, diversity, and difficulty, and lack corresponding frame data. To circumvent these issues, we designed our dataset with the following considerations: 
\textbf{(1) Large-scale:} It is well-known that large-scale datasets can facilitate the development of more complex models. Our dataset surpasses previously commonly used datasets in scale. It comprises 5418 samples recorded by 43 subjects, each lasting about 6 seconds, totaling over 9 hours of recording. 
\textbf{(2) Multi-illumination:} Compared to conventional frame cameras, event cameras possess a higher dynamic range and perform well under low-light conditions. Introducing a variety of illumination conditions increases data diversity and promotes the exploration of modality fusion. We have designed three lighting conditions: bright LED light, dim LED light, and natural light. 
\textbf{(3) Multiple positions:} The distance between the subjects and the camera results in different event rates and spatial distributions, contributing to sample diversity. 
\textbf{(4) Dual-modality:} Most existing datasets are uni-modal, lacking a corresponding color-frame based baseline for comparison and inhibiting the development of SNN-based fusion techniques. Hence, we opted for the DAVIS346 camera, capable of synchronously outputting both event streams and frame streams. A comparison of the parameters of commonly used neuromorphic cameras is shown in Table \ref{table_event_camera}.

\begin{table*}
\centering
\caption{Comparison of Commonly Used Event Cameras}\label{table_event_camera}
\begin{tabular}{llccccc}
\toprule
Camera & Reference  & Resolution & Latency ($\mu s$) & Dynamic range (dB) & Dual-modality & Color frame \\
\midrule
DVS128 & \citep{lich2008} & $128 \times 128$ & 12 & 120 & no & no \\
ATIS & \citep{posch2010qvga} & $240 \times 304$ & 3 & 143 & yes & no \\
DAVIS346 & \citep{inivation2019davis346} & $260 \times 346$ & 20 & 120 & yes & yes \\
\bottomrule
\end{tabular}
\end{table*}

\subsection{Dataset Recording}
\begin{figure}[htb]
    \centering
    \includegraphics[page=1, scale=0.35]{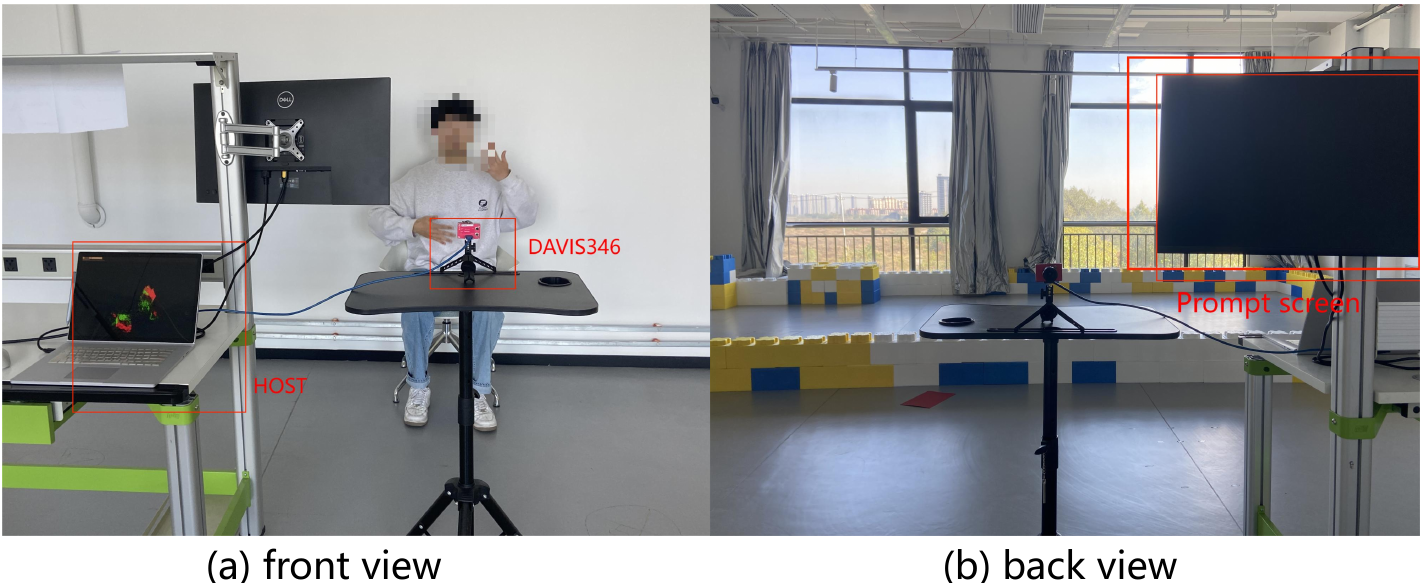}
    \caption{The recording environment. }
    \label{recording_env}
\end{figure}

Figure \ref{recording_env} shows the recording environment. 
The DAVIS346 camera was fixed on the desk, facing the subjects. 
Each subject was required to record six sets of data under three illumination conditions (natural light, bright LED light, dim LED light) and from two positions (front, back). 
Before recording each set of data, the staff would adjust the lighting and the position of the seat. 
The positions "front" and "back" are approximately $1.5m$ and $1m$ from the camera, respectively. 
Under natural light conditions, the curtains were fully opened, resulting in light intensities ranging from 300 to 850 lux, varying with the time of day and weather. For the LED light scenarios, the curtains were drawn closed, and we employed high-power LED strips and low-power LED bulbs to create bright and dim lighting environments, with respective light intensities of about 300lx and 150lx.

Although the DAVIS346 camera is capable of achieving frame rates of up to 40fps, such a short exposure time in dim LED light conditions causes the APS(Active Pixel Sensor) to produce almost black images. Therefore, we set a consistent frame rate of 20fps, which produces a color frame every 0.05 seconds. To collect data from the DAVIS346 camera, we initiated two ROS (Robot Operating System) nodes, one for events and one for frames. These nodes process data from the camera and broadcast it over designated ROS topics. With the \emph{rosbag} utility in ROS, we recorded these topic messages into a \emph{.bag} file. Considering the variety of actions, it was unrealistic to expect each subject to memorize them all. To assist, our system displayed a 3-second instructional video on a prompt screen (see Figure \ref{recording_env}) before recording each action, which the subjects were to mimic. After the instructional video concluded, the recording session began, continuing until the start of the next instructional video.

We selected 20 standard actions from Chinese Sign Language, including \emph{you, I, everybody, at, go, want, listen, talk, see, run, jump, study, eat, drink, wash hands, shower, rest, sleep, home,} and \emph{school}. These actions encompass various motion trajectories and hand postures. Compared to other action recognition datasets, our actions are semantically rich and hence more complex. We also introduced an \emph{OTHER} category for gestures improvised by the subjects, enhancing sample diversity. To ensure diversity in the dataset, we imposed no specific requirements on the speed of the action, the participant's dominant hand, sitting posture, clothing, or hand positioning. For further details, please visit our \href{https://github.com/JasonKitty/DVS-SLR}{webpage}.

\subsection{Dataset Analysis}

\begin{figure}[htb]
    \centering
    \includegraphics[page=1, scale=0.4]{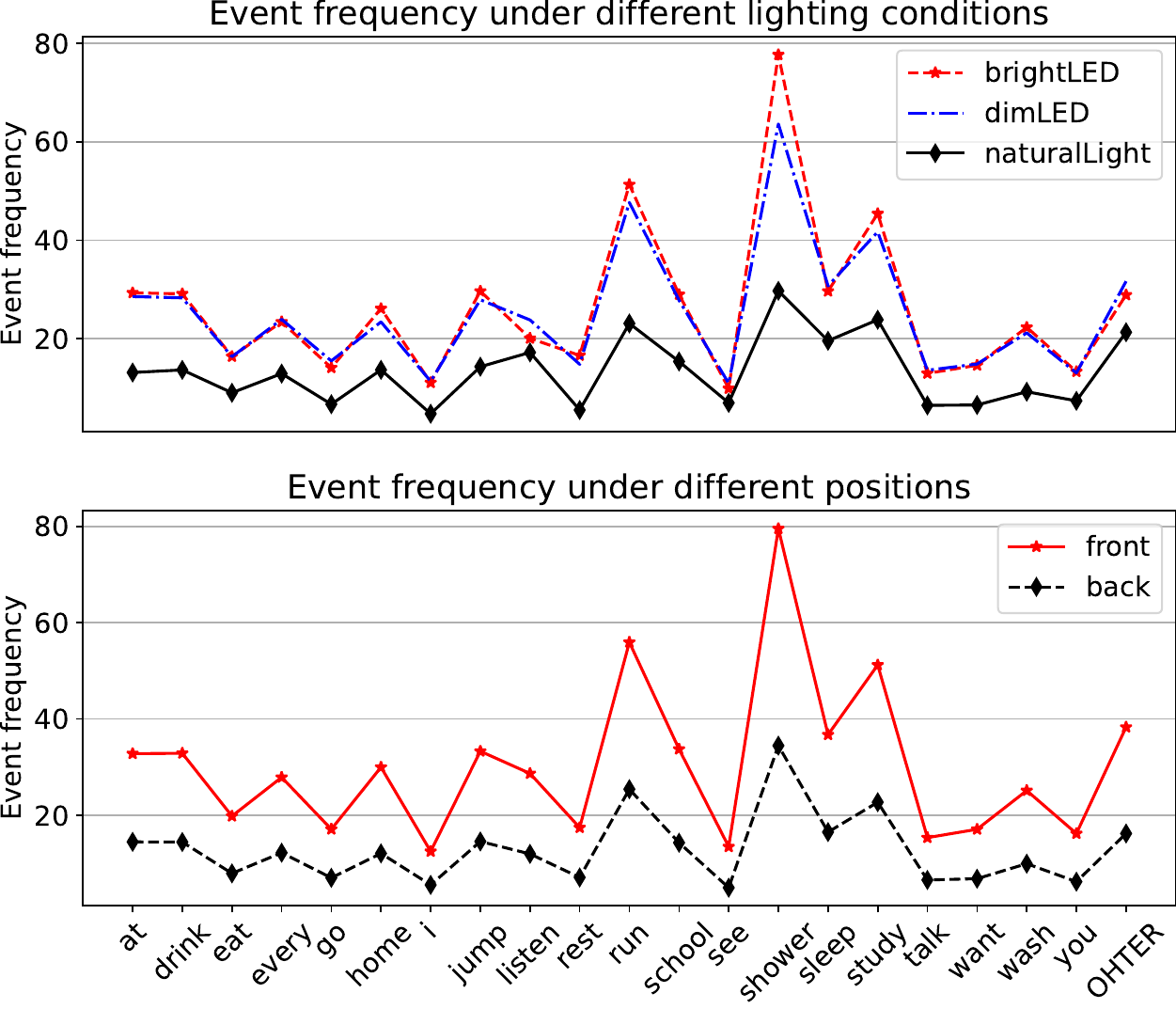}
    \caption{Event frequency of each category in different scenarios. Event frequency refers to the number of events per second, measured in 1,000 events/s.}
    \label{event_frequency}
\end{figure}

To input asynchronous event streams into SNNs, most studies use a method based on accumulating events over fixed time intervals into frames. A rich distribution of event frequencies helps increase sample diversity and enhance the model's generalization performance.

\begin{figure}[htb]
    \centering
    \includegraphics[page=1, scale=0.4]{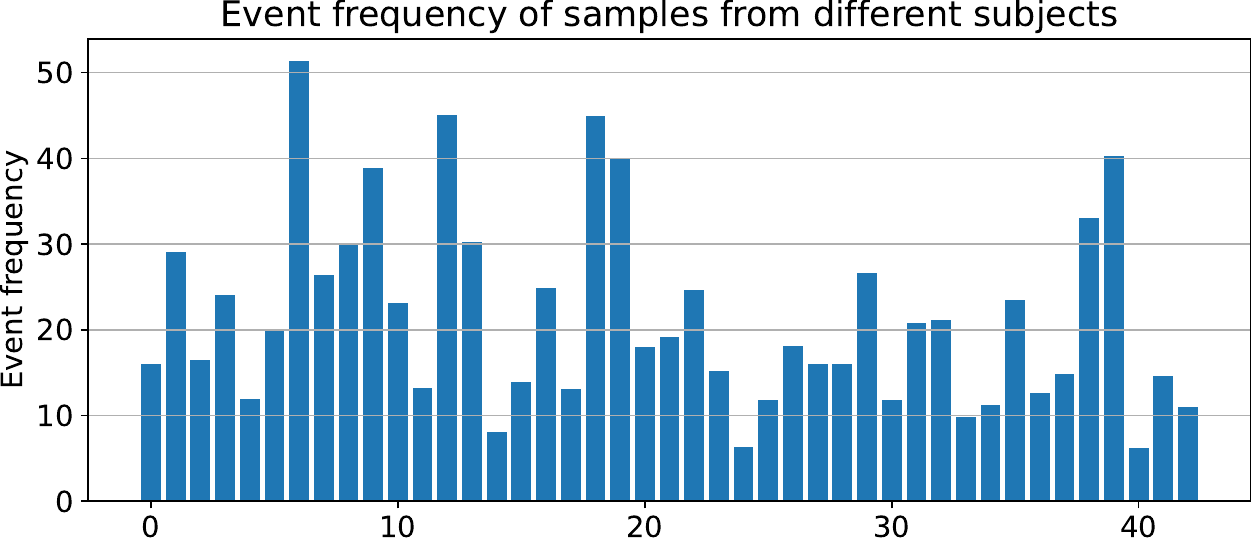}
    \caption{Event frequency distribution of samples from different subjects. Event frequency is measured in 1,000 events/s.}
    \label{event_frequency_sub}
\end{figure}

\begin{figure}[htb]
    \centering
    \includegraphics[page=1, scale=0.37]{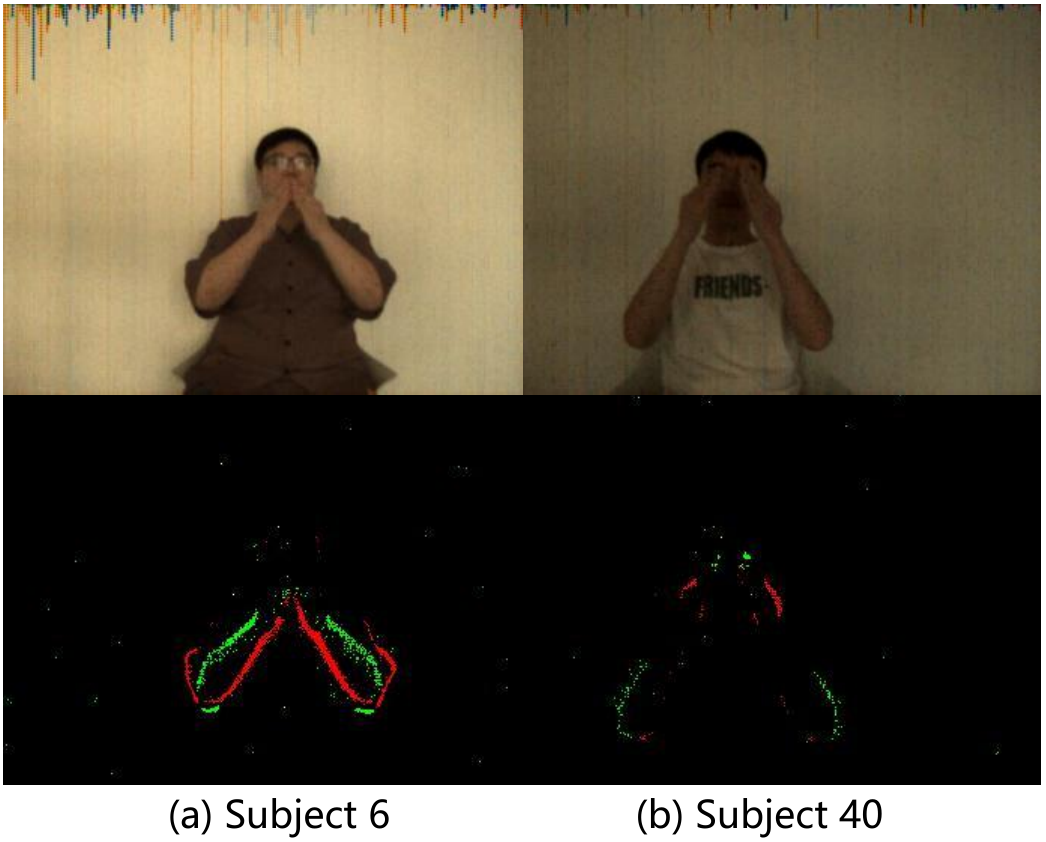}
\caption{Sample comparison from different subjects. (a) Sample from subject 6, with a higher event frequency. (b) Sample from subject 40, with a lower event frequency.}
    \label{user_sample}
\end{figure}

To enhance the diversity of our dataset, we designed various scenarios. Below, we analyze the diversity introduced by these designs. As shown in Figure \ref{event_frequency}, the event frequency under bright LED and dim LED lighting conditions does not differ significantly, while it decreases under natural light. This is partly because the relative change in light intensity is less pronounced under natural light than under LED light, resulting in fewer events captured by the event camera. Additionally, the imperceptible flickering of LED lights leads to a substantial number of events captured. The event frequency is higher in the "front" position than in the "back" position because the subject occupies more pixels in the front position. Furthermore, as seen in Figure \ref{event_frequency_sub}, there is considerable variation in the event frequency of samples produced by different subjects. The highest event frequency is from subject 6, and the lowest is from subject 40, attributed to the higher contrast between subject 6's clothing and the background (see Figure \ref{user_sample}), which triggers more events.

\begin{figure*}[htb]
    \centering
    \includegraphics[page=1, scale=0.63]{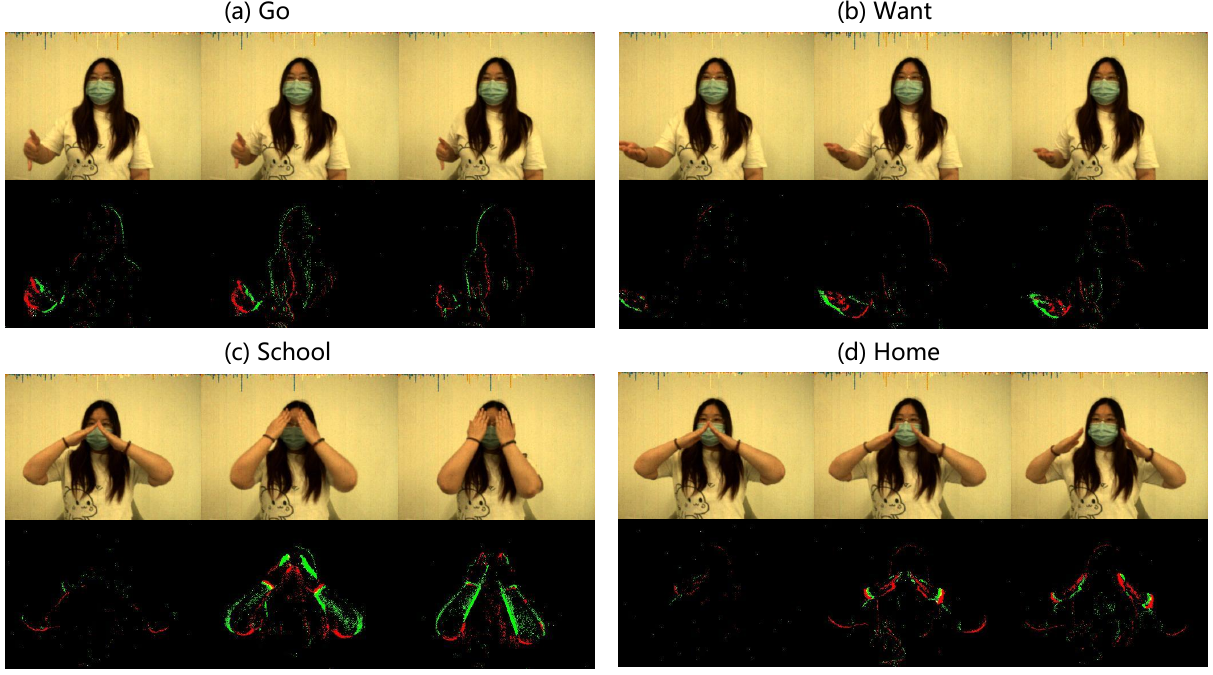}
    \caption{Comparison of different actions. (a) and (b) have the same hand motion trajectory but different hand postures. (c) and (d) have the same hand posture, forming a "roof-like" shape with both hands, but different motion trajectories.}
    \label{recording_comparation}
\end{figure*}

Some of the selected sign language actions are prone to confusion. Some actions share similar trajectories but differ in hand postures (e.g., "want" vs. "go"), while others differ in trajectories but have similar hand postures (e.g., "school" vs. "home"). Figure \ref{recording_comparation} illustrates these confusing similarities.

\begin{table*}
\centering
\caption{Comparison of Neuromorphic Datasets for Action Recognition. Abbreviations: SS - Sample Size, NC - Number of Classes, AD - Average Duration, TRT - Total Recording Time, ML - Multi-light, MP - Multi-position, DM - Dual-Modality}
\begin{tabular}{@{} llcccccccc @{}}
\toprule
Dataset & Reference & Resolution & SS & NC & AD & TRT & ML & MP & DM\\
\midrule
DVS-Gesture & \citep{amir2017low} & $128\times128$ & 1342 & 11 & 5s & 1.86h & $\times$ & $\times$ & $\times$\\
SL-Animal-DVS & \citep{RN68} & $128\times128$ & 1121 & 19 & 4.5s & 1.4h & $\times$ & $\times$ & $\times$\\
ActionRecognition & \citep{RN39} & $260\times346$ & 291 & 10 & 5s & 0.4h & $\times$ & \checkmark & $\times$\\
DailyAction & \citep{liu2021event} & $260\times346$ & 1440 & 12 & 5s & 2h & $\times$ & \checkmark & $\times$\\
\midrule
DVS-SLR (Ours) & - & $260\times346$ & 5418 & 21 & 6s & 9.03h & \checkmark & \checkmark & \checkmark\\
\bottomrule
\label{table_datasets}
\end{tabular}
\end{table*}

Table \ref{table_datasets} compares our dataset with other commonly used neuromorphic datasets for action recognition. Our DVS-SLR dataset has the advantages of larger scale, multiple scenarios, and dual-modality. We anticipate that this dataset will pave the way for fair evaluation of SNNs and the development of SNN-based fusion methods.

\section{Methods}
\subsection{Data Preprocessing}
\label{section_data_representation}
\noindent
\textbf{Data Preprocessing of Event Modality.} 
Event cameras generate an asynchronous stream of events, which is not directly compatible with neural network inputs. Typically, these events are processed by being accumulated into frames. Consider a segment of an event stream, where each event \(e\) is characterized by (\( x, y, t, p \)), with \( x, y \) indicating the spatial coordinates of the event, \( t \) its timestamp, and \( p \) its polarity. Polarity is usually represented as +1 (signifying an increase in brightness) or -1 (indicating a decrease in brightness). The construction of the accumulated frame is as follows:

\begin{align}
I_0(x,y) &= \sum_{e \in E_0} \delta(p - 1) \cdot \delta(x - x_e) \cdot \delta(y - y_e), \\
I_1(x,y) &= \sum_{e \in E_1} \delta(p + 1) \cdot \delta(x - x_e) \cdot \delta(y - y_e),
\end{align}
\noindent
where \(I\in{\mathbb{R}^{2 \times H \times W}}\) denotes the derived accumulated frame, consisting of two channels: \( E_0 \) for frames crafted from positive polarity events and \( E_1 \) for negative polarity events. The function \( \delta(\cdot) \) is the Dirac delta function, which returns 1 when its arguments match and 0 otherwise.
\begin{figure}[htb]
    \centering
    \includegraphics[scale=0.37]{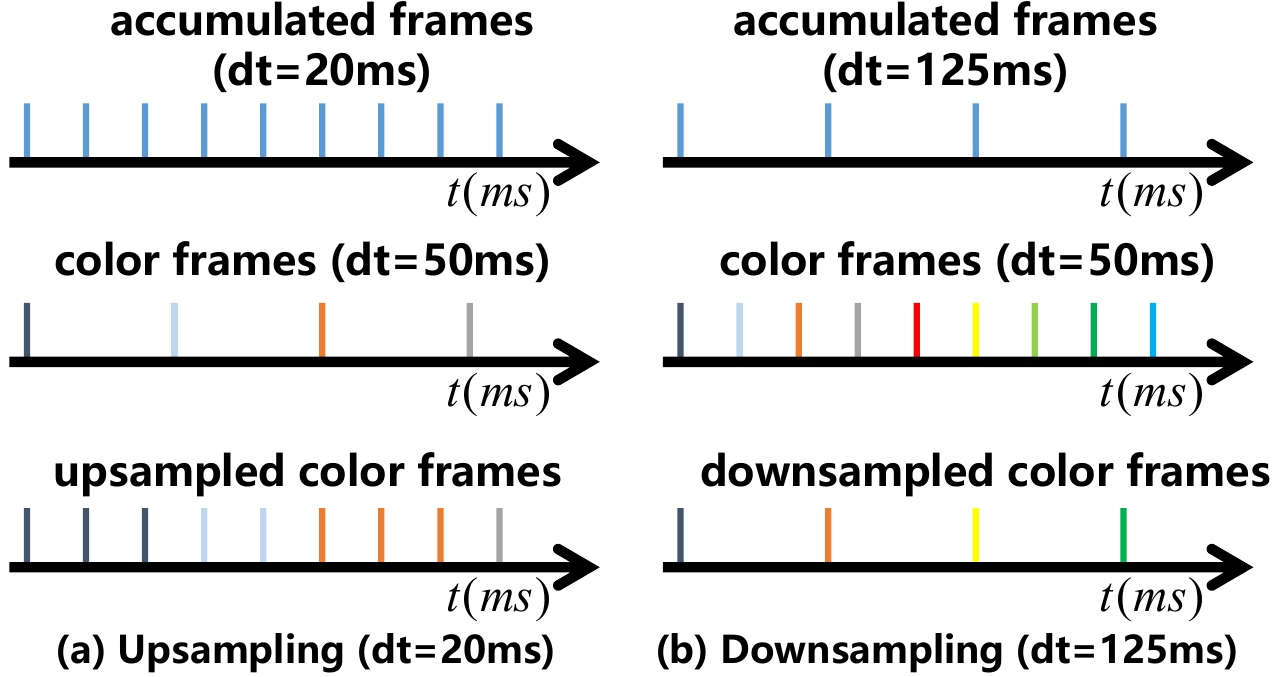}
    \caption{The process of temporal sampling. (a) When $dt$ surpasses the frame interval, the frame stream undergoes upsampling, utilizing adjacent frames to fill the gaps. (b) Conversely, if $dt$ is lower than the frame interval, the frame stream is subjected to downsampling, preserving only the most recent frames.}
    \label{sampling}
\end{figure}

\noindent
\textbf{Dual-Modality Alignment and Sampling.} Our dataset is dual-modal, with the frame modality's frame interval set at 50ms during recording. Since the temporal resolution of the event stream is much higher than that of the frame stream, a smaller time window (e.g., $dt$ < 50ms) is often chosen when accumulating the event stream into frames, leading to asynchrony between the two modalities. To synchronize both modalities, we sample the color frames. As depicted in Figure \ref{sampling}, when $dt$ < 50$ms$, the module retains only the latest color frame, discarding prior ones. Conversely, if $dt$ > 50$ms$, the module fills missing frames using the most recent color frame.

In this study, we employ the LIF neuron model for neuronal modeling due to its optimal balance between precision and computational complexity. Specifically, its dynamics are given by:
\begin{equation}
\label{eq_lif}
\left\{ \begin{gathered}
  \tau \frac{{du}}{{dt}} =  - u + I,u < {V_{th}} \hfill \\
  {\text{fire a spike \&  }}u \leftarrow {u_{reset}},{\text{ }}u \geqslant {V_{th}} \hfill \\ 
\end{gathered}  \right.
\end{equation}
where $u$ is the membrane potential, $\tau$ is the time constant, and $I$ is the input synaptic current. 
Given that the above equation is continuous, for its application in deep learning frameworks, we leverage the Euler formula for its expansion and simplification: 
\begin{equation}
u^{t+1}=(\frac{\tau-1}{\tau})u^{t}+\frac{1}{\tau}I.
\end{equation}

\noindent
The pre-synaptic input $I$ is then approximated through a linear summation of the pre-synaptic spikes \citep{wu2019direct}: 
\begin{equation}
u^{t+1}=(\frac{\tau-1}{\tau})u^{t}+\sum_{j}W_{j}o(j),
\end{equation}
\noindent
where $j$ indicates the index of pre-synapse.
Incorporating the firing-and-resetting mechanism delineated in Eq. (\ref{eq_lif}), we have:

\begin{equation}  
\begin{array}{l}  
  u^{t + 1,n + 1}(i) = (\frac{\tau-1}{\tau})u^{t,n + 1}(i)(1 - o^{t,n + 1}(i)) + \sum\limits_{j = 1}^{l(n)} w_{ij}^n o^{t + 1,n}(j), \\  
  o^{t + 1,n + 1}(i) = f(u^{t + 1,n + 1}(i) - V_{th}), \\   
\end{array}
\label{snn_equ}
\end{equation}
where $n$ and $l(n)$ signify the layer index and its corresponding neuron count. The synaptic weight between the $j$-th neuron of the preceding layer $(n)$ and the $i$-th neuron of the subsequent layer $(n + 1)$ is denoted by $w_{ij}$. The function \( f(\cdot) \) represents the Heaviside function: when \( x < 0 \), \( f(x) = 0 \); otherwise, \( f(x) = 1 \).

\subsection{Baseline Model}
\label{section_baseline_model}
\begin{figure*}[htb]
    \centering
    \includegraphics[page=1, scale=0.27]{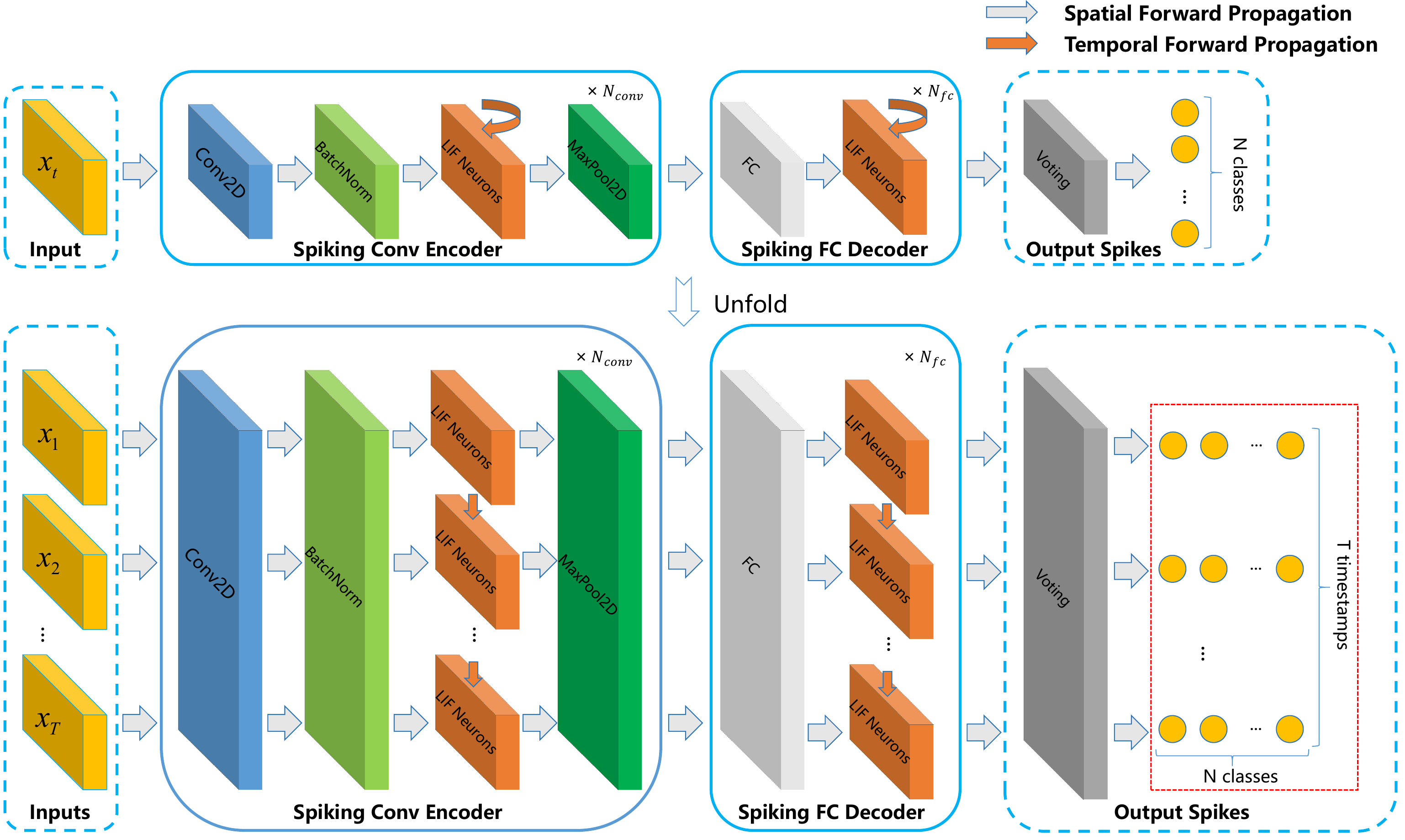}
    \caption{The architecture of our uni-modality baseline model. This model comprises the Spiking Conv Encoder, Spiking FC Decoder, and Voting Module. Unlike traditional CNNs, information in SNNs is transmitted along both the temporal and spatial dimensions.}
    \label{baseline_model}
\end{figure*}
For a novel dataset, a simple yet robust baseline model serves to provide subsequent researchers with a performance yardstick, motivating them to devise superior methodologies to surpass this benchmark. Inspired by the architecture proposed in \citep{fang2021incorporating}, our baseline model comprises three key components: the Spiking Conv Encoder, the Spiking FC (Fully Connected) Decoder, and the Output module (see Figure \ref{baseline_model}).

The Spiking Conv Encoder consists of $N_{conv}$ sets of Conv2D-BN (Batch Normalization)-LIF (Leaky Integrate-and-Fire) -Maxpool2D modules. All Conv2D layers have a kernel size of 3, stride of 1, and padding of 1. Each pooling layer employs a kernel size of 2 and a stride size of 2.

The Spiking FC Decoder is composed of $N_{fc}$ sets of FC-Dropout-LIF modules. The last FC layer has \( N \) neurons, where \( N \) should be a multiple of the class count \( C \). The sparse output spikes are then directed into a voting module to produce the final output.

The Voting Module is designed to relieve the challenges posed by the sparse nature of spikes. For each timestep, it processes the incoming $N$ spikes, dividing them along the spatial dimension into $C$ groups, and subsequently computes the average value for each group as the final output of the current timestamp. 

\subsection{Fusion Strategy}
\label{Complementarity}
\begin{figure}[htb]
    \centering
    \includegraphics[scale=0.23]{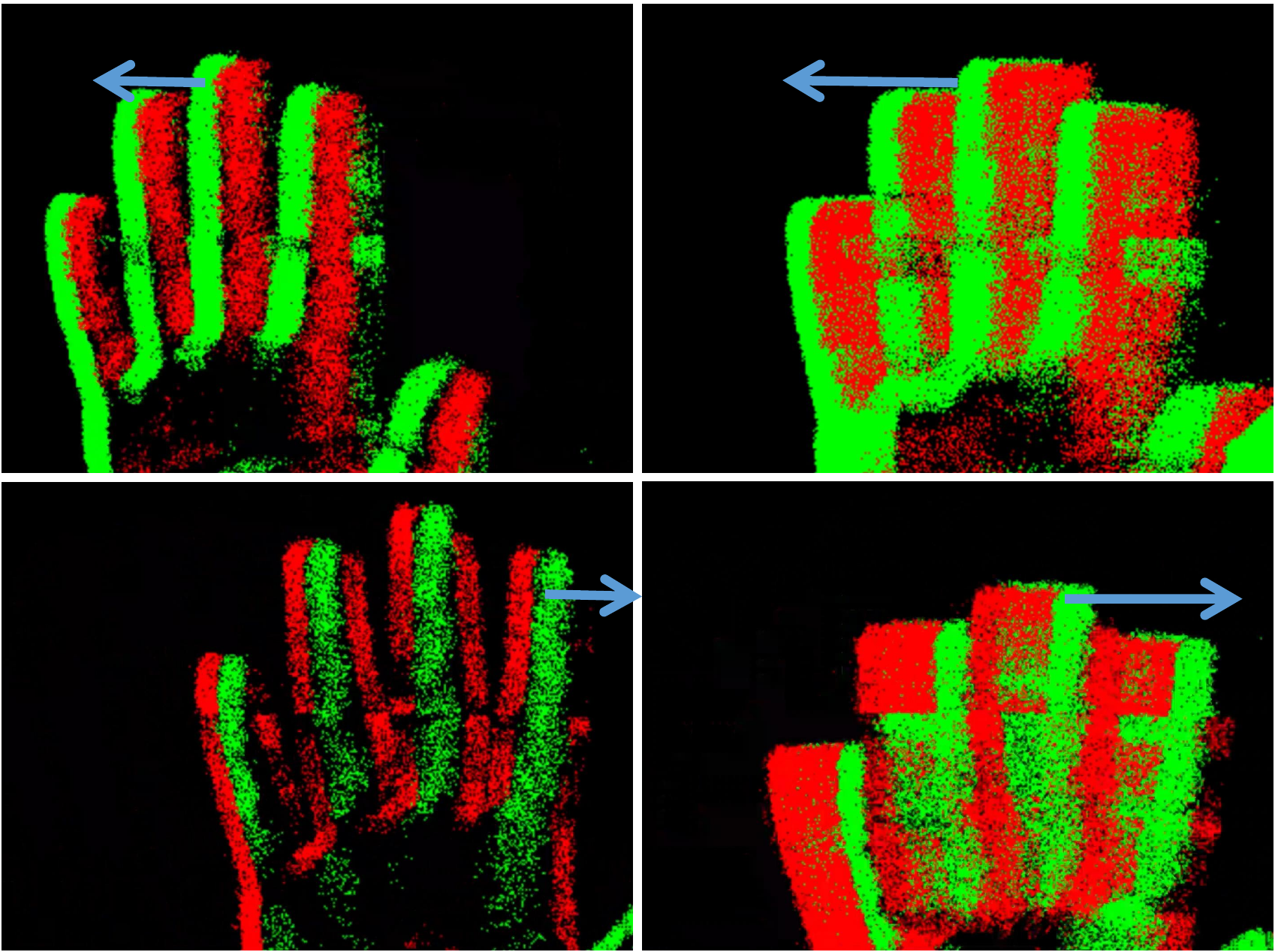}
\caption{Illustration of Accumulated Frame characteristics. \textbf{Top-left:} Slow hand wave from right to left; \textbf{Top-right:} Fast hand wave from right to left; \textbf{Bottom-left:} Slow hand wave from left to right; \textbf{Bottom-right:} Fast hand wave from left to right. In this illustration, green and red pixels indicate negative and positive event polarities, respectively. Motion vectors are annotated within the images for reference. During movement, negative events come before positive ones. This pattern hints at the direction of movement. Also, faster movements create more events within a given duration (dt), suggesting that the event frequency conveys, in part, the speed of the movement.}
    \label{motion_sample}
\end{figure}

\begin{figure}[htb]
    \centering
    \includegraphics[page=2, scale=0.36]{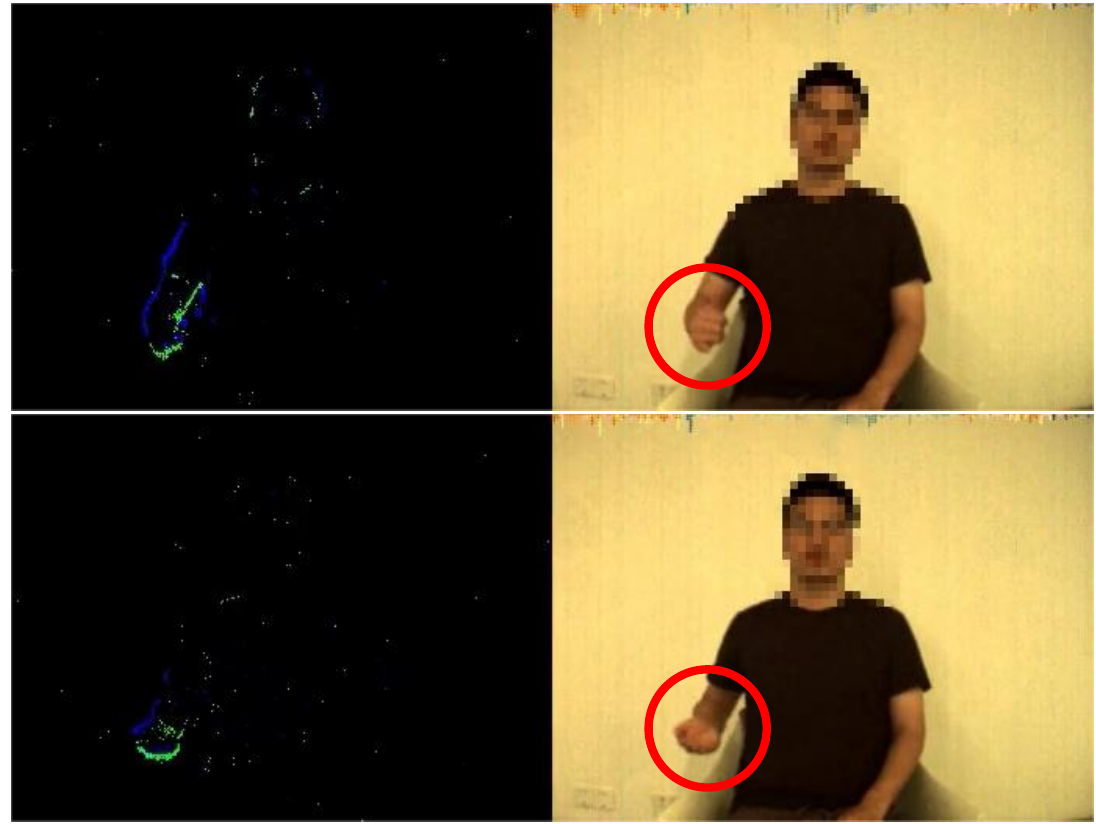}
\caption{Illustration of Color Frame Characteristics. The action shown at the top signifies "go", while the one at the bottom represents "want". Despite both actions exhibiting a similar cyclical arm movement from the front to the back, they differ in hand gestures. In instances like these, the frame modality, compared to the event modality, offers richer texture details, enhancing the extraction of discriminative features.}
    \label{texture_sample}
\end{figure}

\subsubsection{Analysis of Modality Complementarity}
Effective action recognition systems integrate both temporal and spatial features, harnessing the strengths of each to build robust models. Modern deep learning approaches, especially those leveraging two-stream architectures \citep{simonyan2014two, carreira2017quo, feichtenhofer2019slowfast}, have shown good performance. We draw inspiration from these successful approaches in two main aspects:
\begin{enumerate}[(1)] 
\item
Two-stream architectures employ convolutional kernels that extend into the temporal dimension. This allows the convolutional networks to effectively extract temporal features. In contrast, our approach leverages the intrinsic temporal modeling of SNNs, eliminating the need for 3D convolution operations.
\item
Two-stream architectures utilize inputs that are complementary. Apart from the traditional frame stream, they introduce optical flow into the system. However, its computational demand often limits its use in real-time applications. In our design, we view the event modality and the frame modality as complementary. The event stream implicitly reflects the motion direction and speed of the object (see Figure \ref{motion_sample}). On the other hand, the frame stream captures detailed spatial features, such as shape, texture, and color of objects (see Figure \ref{texture_sample}). 
\end{enumerate}
Nevertheless, seamlessly integrating both within the SNNs framework remains a challenging endeavor.

\subsubsection{Cross-modality Attention}
\begin{figure*}[htb]
    \centering
    \includegraphics[page=2,width=\textwidth]{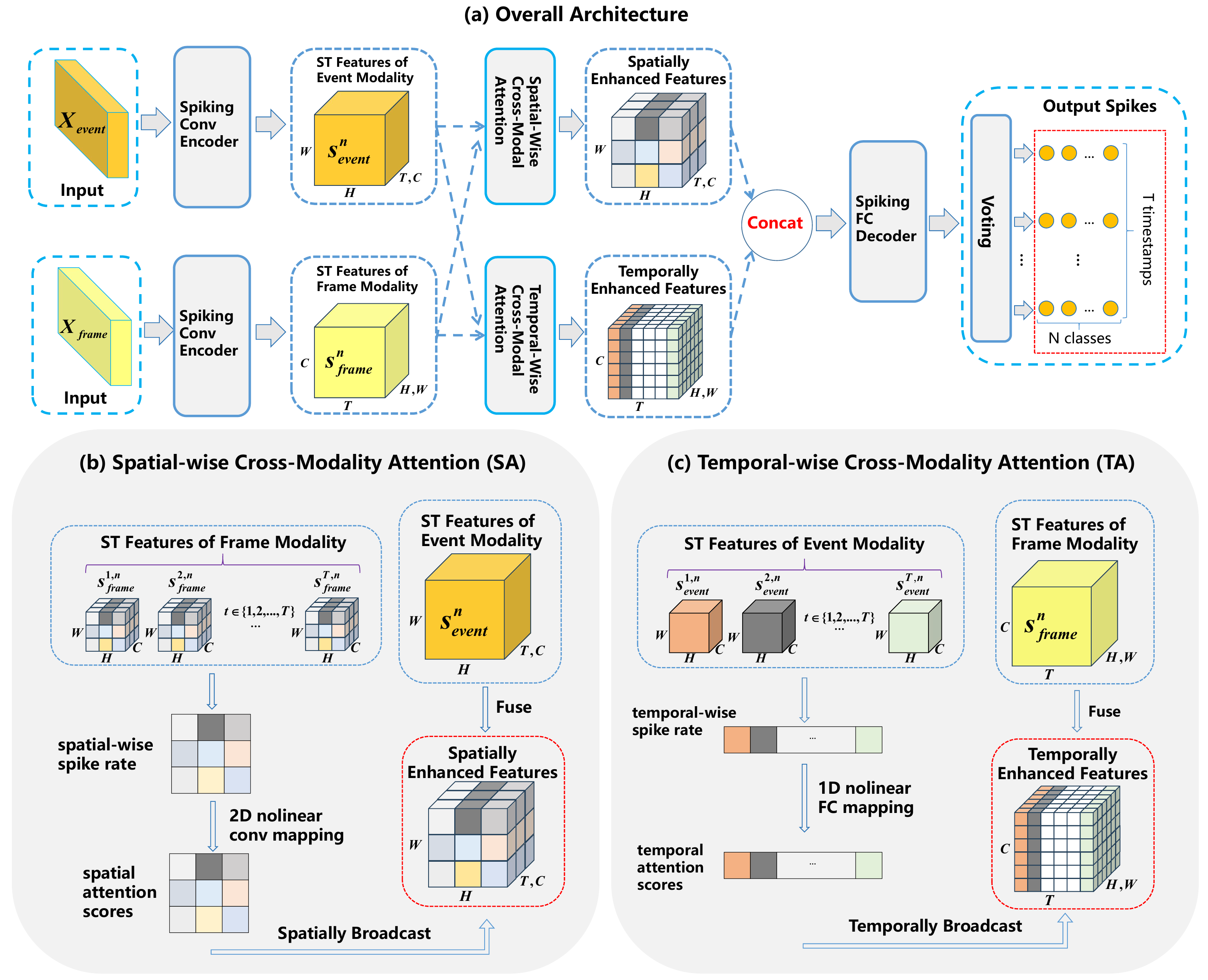}
\caption{Schematic of the cross-modality fusion based action recognition framework. (a) depicts the overall framework, where the Spiking Conv Encoder, Spiking FC Decoder, and Output module all originate from the baseline model. (b) and (c) detail the spatial-wise and temporal-wise cross-modality attention module, respectively.}
    \label{overall_model}
\end{figure*}
Drawing from our analysis, event streams excel at capturing dynamic changes, whereas frame streams provide more extensive spatial information. Embracing these distinct characteristics, our fusion approach seeks to extract temporal attention scores from the event modality and spatial attention scores from the frame modality, subsequently cross-fusing them to enhance the synergy between the two. Inspired by the Convolutional Block Attention Module (CBAM) \citep{woo2018cbam} and Attention Spiking Neural Networks \citep{yao2023atten}, we employ Temporal Attention (TA) and Spatial Attention (SA) to generate attention scores. Intuitively, TA instructs the network on when to pay attention, while SA indicates where to pay attention. Figure \ref{overall_model} visually represents this process. The spike firing rate reflects the neurons' activity intensity and can be interpreted as the degree of response to specific information \citep{salinas2000impact}. To augment the learning of attention scores, unlike \citep{yao2023atten}, which uses the membrane potential as the input, we employ the spike firing rate as the input to our attention module.

Formally, let \(s_e^n \in {\mathbb{R}^{T \times C \times H_n \times W_n}}\) and \(s_f^n \in {\mathbb{R}^{T \times C \times H_n \times W_n}}\) denote the spatio-temporal features of the event and frame modalities, respectively. Here, \( n \) is the layer index for the final convolutional block, \( T \) represents the timesteps, and \( H_n \), \( W_n \), and \( C \) are the height, width, and channel dimensions of the feature map at layer \( n \).

For the event modality, the temporal spike firing rate is defined as \(r_e^n = [ \cdots ,r_e^{t,n} \cdots ] \in {\mathbb{R}^{T}}\), and can be computed for each timestep as:
\begin{equation}\label{3.4.2.eq1}
r_e^{t,n} = \frac{1}{{C \times {H_n} \times {W_n}}}\sum\limits_k^C {\sum\limits_i^{{H_n}} {\sum\limits_j^{{W_n}} {s_e^{t,n}(k,i,j)} } }.
\end{equation}
In parallel, the spatial spike firing rate for the frame modality is denoted by \(r_f^n = {\left\{ {r_f^n(i,j)} \right\}_{{H_n} \times {W_n}}} \in {\mathbb{R}^{{H_n} \times {W_n}}}\), and for each spatial position, it's computed as:
\begin{equation}\label{3.4.2.eq2}
r_f^{n}(i,j) = \frac{1}{{T \times C}}\sum\limits_t^T {\sum\limits_k^C {s_f^n(t,k,i,j)} }.
\end{equation}
\noindent
Then, the temporal attention scores \({d_e^n} \in {\mathbb{R}^{T}}\) are learned through two nonlinear FC mappings:
\begin{equation}\label{3.4.2.eq3}
d_e^n = \sigma (M_2^n\delta (M_1^nr_e^n)),
\end{equation}
where \(M_1^n \in {\mathbb{R}^{T \times \frac{T}{r}}}\) and \(M_2^n \in {\mathbb{R}^{\frac{T}{r} \times T}}\) are learnable parameters, \(\delta\) and \(\sigma\) represent the sigmoid and ReLU function.

\noindent
Similarly, the spatial attention scores, denoted as \({d_f^n} \in {\mathbb{R}^{{H_n} \times {W_n}}}\), are learned through a convolutional mapping:
\begin{equation}\label{3.4.2.eq4}
d_f^n = \sigma(M_3^n * r_f^n),
\end{equation}
where the \(M_3^n \in {\mathbb{R}^{3 \times 3}}\) signifies a learnable convolution kernel,  \(\sigma\) represents the ReLU function, and \(*\) indicates the convolution operation.
Subsequently, temporal attention scores and frame features are cross-fused to yield temporally augmented features, \(\tilde s_f^n = [ \cdots ,\tilde s_f^{t,n} \cdots ] \in {\mathbb{R}^{T \times C \times {H_n} \times {W_n}}}\). Explicitly, the fusion for each timestep is given by:
\begin{equation}\label{3.4.2.eq5}
\tilde s_f^{t,n} = d_e^{t,n} \times s_f^{t,n}.
\end{equation}
In a mirrored operation, spatial attention scores and event features are cross-fused, resulting in spatially augmented features, \(\tilde s_e^n \in {\mathbb{R}^{T \times C \times {H_n} \times {W_n}}}\). For each spatial coordinate, the fusion is:
\begin{equation}\label{3.4.2.eq6}
\tilde s_e^n(:,:,i,j) = d_f^n(i,j) \times s_e^n(:,:,i,j).
\end{equation}

\noindent
Finally, the cross-modality augmented features are concatenated, yielding the final CMA module output:
\begin{equation}\label{3.4.2.eq7}
{{\tilde s}^n} = [\tilde s_e^n;\tilde s_f^n] \in {\mathbb{R}^{T \times 2C \times {H_n} \times {W_n}}}.
\end{equation}

In order to make it clear to the reader, we have also presented the computation processes for TA and SA modules in pytorch pseudocode form, as illustrated in Algorithm \ref{alg:temporal_attention} and \ref{alg:spatial_attention}.

\begin{algorithm}[htb]
\caption{Temporal Attention Fusion}
\label{alg:temporal_attention}
\begin{algorithmic}[1]
\State \textbf{Input:}
\State \quad $s_e[T, C, H, W]$: Event modality features
\State \quad $s_f[T, C, H, W]$: Frame modality features
\State
\State \textbf{Output:}
\State \quad $TempEnh[T, C, H, W]$: Temporally enhanced features
\State
\State \textbf{Begin:}
\State // Compute Temporal Spike Firing Rate for event modality
\State $r_e \gets \text{mean}(s_e, \text{axis}=(C, H, W))$  // Dimensions: [T]
\State
\State // Compute Temporal Attention Scores (TA)
\State $TA \gets \sigma(M_2^n \cdot \delta(M_1^n \cdot r_e))$  // Dimensions: [T]
\State
\State // Apply Temporal Attention to frame features using broadcasting and multiplication
\State // Dimensions: [T, C, H, W]
\State $TempEnh \gets TA.unsqueeze(1).unsqueeze(2).unsqueeze(3) \cdot s_f$
\State \textbf{End}
\end{algorithmic}
\end{algorithm}

\begin{algorithm}
\caption{Spatial Attention Fusion}
\label{alg:spatial_attention}
\begin{algorithmic}[1]
\State \textbf{Input:}
\State \quad $s_e[T, C, H, W]$: Event modality features
\State \quad $s_f[T, C, H, W]$: Frame modality features
\State
\State \textbf{Output:}
\State \quad $SpaEnh[T, C, H, W]$: Spatially enhanced features
\State
\State \textbf{Begin:}
\State // Compute Spatial Spike Firing Rate for frame modality
\State $r_f \gets \text{mean}(s_f, \text{axis}=(T, C))$  // Dimensions: [H, W]
\State
\State // Compute Spatial Attention Scores (SA)
\State $SA \gets \sigma(\text{Convolve}(M_3, r_f))$  // Dimensions: [H, W]
\State
\State // Apply Spatial Attention to event features using broadcasting and multiplication
\State // Dimensions: [T, C, H, W]
\State $SpaEnh \gets SA.unsqueeze(0).unsqueeze(0) \cdot s_e$
\State \textbf{End}
\end{algorithmic}
\end{algorithm}

\subsection{Overall Training}
The overall architecture is illustrated in Figure \ref{overall_model}. Its fundamental components are the same as the baseline model, but with the addition of dual-modality input and CMA module. The model receives synchronized dual-modality inputs, \(X_{event} \in {\mathbb{R}^{T \times 2 \times H \times W}}\) and \(X_{frame} \in {\mathbb{R}^{T \times 3 \times H \times W}}\). Each modality's data is input into the Spiking Conv Encoder to extract spatio-temporal features. These features are then fed into the spatial-wise CMA and temporal-wise CMA for cross-modality fusion. The fused features are concatenated and fed into the Spiking FC Decoder and Voting Module to produce the output spikes.

Formally, let \(T\) denote the timesteps, and \(C\) represent the number of classes. For each input, the network produces an output \(O = {\left\{ {o_i^t} \right\}_{T \times C}}\). Ideally, for an input labeled \(l\), the neuron corresponding to the correct class should fire at every timestep, with other neurons remaining inactive. This target behavior can be expressed mathematically as:
\begin{equation}\label{3.5.eq10}
y_i^t = \left\{ \begin{array}{ll}
  1 & \text{if } i = l \\
  0 & \text{if } i \ne l 
\end{array}  \right.
\end{equation}
The discrepancy between the predicted output and this target is captured using the mean squared error (MSE) as loss function:
\begin{equation}\label{3.5.eq1}
L = \frac{1}{{T \times C}}\sum\limits_t {\sum\limits_i {{{(o_i^t - y_i^t)}^2}} }.
\end{equation}
Furthermore, the predicted label \( \hat l \) corresponds to the neuron's index showcasing the utmost spike firing rate, defined as \( \hat l = \arg {\max _i}\sum\limits_t {o_i^t} \).

In this study, we utilize STBP \citep{wu2019direct} to train our network. The gradient is back-propagated simultaneously across both temporally and spatially:

\begin{align}
\frac{{\partial L}}{{\partial o_i^{t,n}}} &= \sum\limits_{j = 1}^{l(n + 1)} {\frac{{\partial L}}{{\partial o_j^{t,n + 1}}}} \frac{{\partial o_j^{t,n + 1}}}{{\partial o_i^{t,n}}} + \frac{{\partial L}}{{\partial o_i^{t + 1,n}}}\frac{{\partial o_i^{t + 1,n}}}{{\partial o_i^{t,n}}}, \label{3.5.eq2} \\
\frac{{\partial L}}{{\partial u_i^{t,n}}} &= \frac{{\partial L}}{{\partial o_i^{t,n}}}\frac{{\partial o_i^{t,n}}}{{\partial u_i^{t,n}}} + \frac{{\partial L}}{{\partial o_i^{t + 1,n}}}\frac{{\partial o_i^{t + 1,n}}}{{\partial u_i^{t,n}}}, \label{3.5.eq3}
\end{align}
where \(l(n+1)\) denotes all the neurons in the \((n+1)\)th layer. 

\begin{figure}[htb]
    \centering
    \includegraphics[scale=0.4]{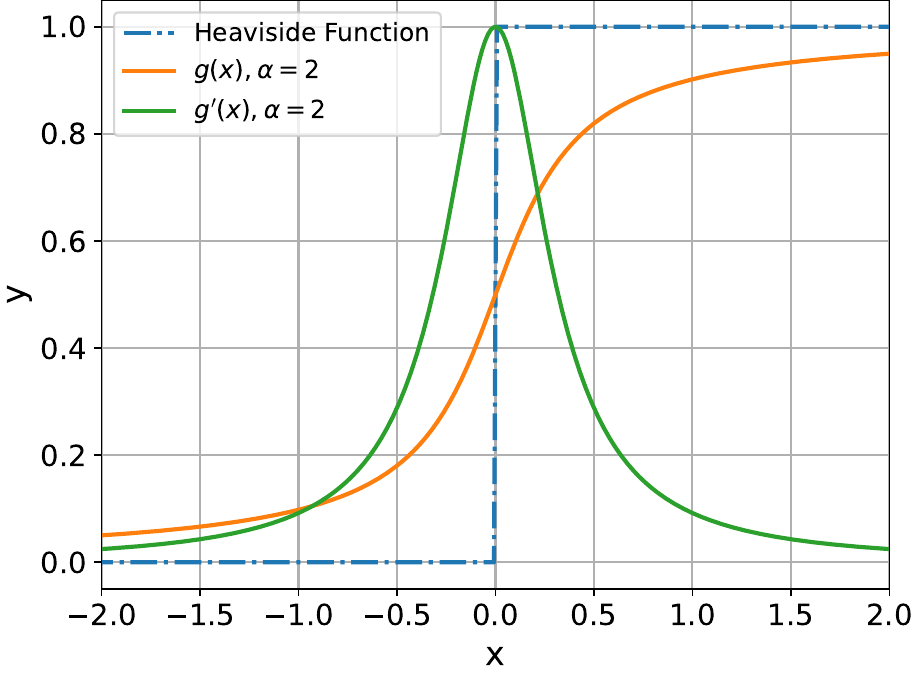}
    \caption{Illustration of the Heaviside function, its approximation \( g(x) \), and the surrogate gradient ${g^\prime }(x)$.}
    \label{surrogate}
\end{figure}

The firing process in SNNs is inherently non-differentiable. We approximate the Heaviside function using:

\begin{equation}\label{3.5.eq4}
g(x) = \frac{1}{\pi }{\text{arctan}}\left(\frac{\pi }{2}\alpha x\right) + \frac{1}{2}
\end{equation}
\noindent
Its derivative or the surrogate gradient can be expressed as:

\begin{equation}\label{3.5.eq5}
g'(x) = \frac{{dg}}{{dx}} = \frac{\alpha }{{2(1 + \left(\frac{\pi }{2}\alpha x\right)^2)}}
\end{equation}
\noindent
Figure \ref{surrogate} provides a visual representation of the Heaviside function, its approximation \( g(x) \), and the surrogate gradient \( {g^\prime }(x) \). Although the surrogate gradient isn't an exact representation, it offers a commendable approximation of the spike firing behavior and its efficacy has been substantiated in numerous studies \citep{zheng2021going, zhang2020temporal, yang2021backpropagated}.

\section{Experiments}
Considering that our work is mainly divided into two aspects, on the one hand, a benchmark dataset DVS-SLR is proposed, which has higher temporal correlation and is more helpful to enable SNNs to exploit spatio-temporal nature. On the other hand, a cross-modality fusion method is proposed, which can improve the synergy of frame modality and event modality.

Therefore, our experiments are also divided into two parts, the first part is to verify that our DVS-SLR dataset has high temporal correlation, while the second part is the dual-modality fusion experiments. We also provide a uni-modality recognition benchmark to offer researchers a reference. Our experiments are implemented based on the SPAIC \citep{hong2024spaic} platform. SPAIC is a dedicated spike-based artificial intelligence computing platform which balances flexibility and efficiency. In addition, we will release a spikingjelly \citep{fang2023spikingjelly} version of the preprocessing program.

\subsection{Temporal Correlation Experiments}
\label{Dataset_Analysis}
As mentioned above, DVS-SLR is a dual-modality dataset. When used solely in its event modality, it serves as an ideal neuromorphic benchmark dataset for evaluating SNNs due to its higher temporal correlation, effectively harnessing SNNs' intrinsic spatio-temporal representation. We share the perspective of \citep{RN29, RN30, RN31} that most of commonly used neuromorphic datasets do not aid in exploring the spatio-temporal properties of SNNs. In this section, we validate this viewpoint through experiments.

\subsubsection{Experimental Design}
The objective of our experiments is to validate the significance of temporal information in neuromorphic data and its influence on the performance of the tested models. \citep{RN29} approached this by comparing the performance of two unsupervised STDP variants - rate-dependent STDP and STDP-tempotron - to assess the importance of spatial and temporal information. However, the performance of unsupervised STDP heavily relies on the hyperparameters settings, and the introduction of two different models could potentially lead to bias. Our strategy, in contrast, employs a consistent SNN model, and analyzes the importance of temporal information by scrambling the spike timings in the original data and completely eliminating temporal information.

\begin{figure}[htb]
    \centering
    \includegraphics[scale=1]{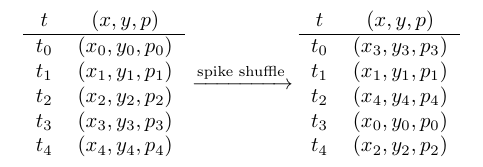}
    \caption{Illustration of the spike timing confusion process, where the order of spikes in the event stream is randomly shuffled.}
    \label{spike_time_confusion}
\end{figure}

\noindent
\textbf{Spike Timing Confusion.} In this experiment, we scramble the spike timing in each event stream during the inference phase, thereby supplying the model with incorrect temporal information (see Figure \ref{spike_time_confusion}). If there is a significant decrease in recognition accuracy, it indicates that precise spike timing plays a critical role, and such datasets can be more helpful in fully leveraging the spatio-temporal property of SNNs.
\begin{figure}[htb]
    \centering
    \includegraphics[scale=0.55]{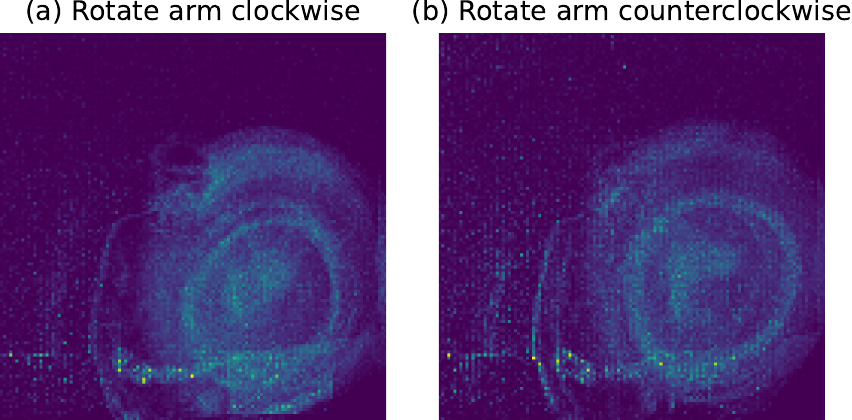}
    \caption{Illustration of temporal information elimination. Relying solely on spatial domain information is insufficient to differentiate between clockwise and counter-clockwise arm rotation movements.}
    \label{Temporal_Elimination}
\end{figure}

\noindent
\textbf{Temporal Information Elimination.} In this experiment, we aim to completely remove the temporal information to observe its impact on model performance. Specifically, the entire event stream is accumulated into one frame (as depicted in Figure \ref{Temporal_Elimination}) for training a SNN. If there is a notable decline in recognition accuracy, it signifies that the temporal information holds a significant role in achieving precise recognition.

\subsubsection{Comparative Datasets}
We compare three of the most commonly used neuromorphic datasets in the SNN domain: DVS-Gesture \citep{amir2017low}, N-MNIST \citep{RN32}, and DVS-CIFAR10 \citep{RN35}. 

\noindent
\textbf{DVS-Gesture} comprises a series of dynamic gestures performed by 29 different individuals, spanning 11 classes, and recorded using DVS cameras under three different lighting conditions. 

\noindent
\textbf{N-MNIST} is generated by moving the ATIS vision sensor across each image in the MNIST database, encompassing all 60,000 training images and 10,000 test images.

\noindent
\textbf{DVS-CIFAR10} transforms the well-known CIFAR-10 dataset into a dynamic version. It contains 1000 samples for each of the 10 categories in CIFAR-10, resulting in a total of 10,000 samples.

\subsubsection{Experimental Settings}
All datasets were preprocessed as described in Section \ref{section_data_representation}. The parameters \(dt\) and \(T\) were set to 10ms and 20, respectively. Training settings and neuron hyperparameters are comprehensively listed in Table \ref{training_param}, and all experiments in this study followed this configuration. Additionally, Table \ref{model_param} provides the model structure parameters for each dataset.

\begin{table}[h]
    \centering
    \begin{tabular}{ccccccc}
        \toprule
        \multicolumn{4}{c}{Training Settings} & \multicolumn{3}{c}{Neuron Settings} \\
        \cmidrule(r){1-4}
        \cmidrule(l){5-7}
        Epoch & \(lr\) & Optimizer & \(\alpha\) & \(V_{th}\)  & \(V_{reset}\) & \(\tau\) \\
        \midrule
        100       & 0.1 & Adam & 2      & 1.0   & 0.0      & 2 \\
        \bottomrule
    \end{tabular}
    \caption{Training settings and neuron hyperparameters. \(\alpha\) corresponds to the surrogate gradient parameter in equation (\ref{3.5.eq5}), and \(\tau\) represents the delay constant of the LIF neuron.}
    \label{training_param}
\end{table}

\begin{table}[htb]
    \centering
    \begin{tabular}{lcc}
        \toprule
        Dataset & $N_{conv}$ & $N_{fc}$  \\
        \midrule
        N-MNIST & 2 & 2 \\
        DVS-CIFAR10 & 4 & 2 \\
        DVS-Gesture & 5 & 2 \\
        DVS-SLR & 5 & 2 \\
        \bottomrule
    \end{tabular}
\caption{Model structure parameters for different datasets. $N_{conv}$ and $N_{fc}$ are defined in Figure \ref{baseline_model}.}
\label{model_param}
\end{table}
\subsubsection{Results \& Discussion}

\begin{table*}[ht]
\centering
\begin{tabular}{lccc}
\hline
Dataset & Baseline & \#Expt1 & \#Expt2 \\
\hline
N-MNIST & 99.21 & 87.71 (-11.58\%) & 97.72 (-1.50\%) \\
DVS-CIFAR10 & 60.51 & 51.50 (-14.86\%) & 57.13 (-5.59\%) \\
DVS-Gesture & 94.78 & 69.07 (-27.15\%) & 83.14 (-12.26\%) \\
DVS-SLR (ours) & 70.49 & 41.54 (-41.08\%) & 51.46 (-26.99\%) \\
\hline
\end{tabular}
\caption{Comparison of recognition accuracy across different experimental conditions. The values in parentheses indicate the percentage decrease in performance compared to the baseline. \#Expt1 refers to the Spike Timing Confusion experiment, while \#Expt2 refers to the Temporal Information Elimination experiment.}
\label{temporal_corr_expt}
\end{table*}

The results presented in Table \ref{temporal_corr_expt} clearly demonstrate that both providing the model with inaccurate temporal information (\#Expt1) and retaining only the spatial information (\#Expt2) lead to a reduction in recognition accuracy. 
This decrease is particularly significant in dynamic datasets like DVS-Gesture and our DVS-SLR, highlighting their rich temporal content. For datasets like N-MNIST and DVS-CIFAR10, which are derived from saccade transformations, completely discarding temporal information—i.e., accumulating it into a single frame—still achieves almost the same level of recognition accuracy. In these cases, the SNN is more like modeling each timestep separately and then voting on the result, which does not exploit the implicit temporal dynamics in the data. 
Our DVS-SLR demonstrates a more marked decline in accuracy in both experiments compared to DVS-Gesture, indicative of a stronger temporal correlation where precise spike timing is vital for accurate recognition. This can be attributed to our deliberate selection of sign language gestures, which accentuates the importance of temporal information. Additionally, the diversity in scenarios, subjects, and gestures in our DVS-SLR results in a lower baseline, offering researchers more room for improvement, as opposed to seeking marginal gains on simpler datasets.

Ultimately, our DVS-SLR is intended to serve as a benchmark dataset to encourage further research within the SNN community, particularly in exploring the spatio-temporal properties of SNNs.

\subsection{Dual-modality based Recognition}
\subsubsection{Experimental Setup}
The dual-modality data were processed into synchronized, continuous inputs \(X_{event} \in {\mathbb{R}^{T \times 2 \times H \times W}}\) and \(X_{frame} \in {\mathbb{R}^{T \times 3 \times H \times W}}\), following the method outlined in Section \ref{section_data_representation}. The training settings and neuron hyperparameters are detailed in Table \ref{training_param}. During both training and inference phases, we randomly extracted segments of length \( t_{lat} \) from the original data, employing these segments as inputs to our model. All experiments were replicated three times, with the mean value reported as the result.

First, we established a uni-modality baseline using the baseline model described in Section \ref{section_baseline_model} to provide a reference point for dual-modality fusion. To thoroughly evaluate our CMA-based fusion method, we conducted extensive experiments with various fusion methods, including performance under different latency configurations and scenarios. Ablation studies were also carried out to assess the effectiveness of Temporal Attention (TA) and Spatial Attention (SA). 

\subsubsection{Evaluation}
\begin{table}[htb]  
    \centering  
    \begin{tabular}{lllcc}  
        \toprule  
        \multirow{2}*{Latency (s)} & \multirow{2}*{\( dt \) (ms)} & \multirow{2}*{\( T \)} & \multicolumn{2}{c}{Modality} \\  
        \cmidrule{4-5}  
          & & & Event & Frame\\  
        \midrule  
        \multirow{3}{*}{0.2} & 10 & 20  & 70.49     & 64.71 \\  
                      & 20 & 10  & 70.78     & 65.09 \\  
                      & 50 & 4   & 69.05     & 65.14 \\  
        \midrule  
        \multirow{3}{*}{0.5} & 10 & 50  & 77.78     & 74.71 \\  
                      & 20 & 25  & 77.43     & 74.43 \\  
                      & 50 & 10  & 77.63     & 75.69 \\  
        \midrule  
        \multirow{3}{*}{1}   & 20 & 50  & 82.32     & 81.71 \\  
                      & 50 & 20  & 81.89     & 80.82 \\  
                      & 100& 10  & 82.54     & 80.96 \\  
        \bottomrule  
    \end{tabular}  
\caption{Uni-modality recognition accuracy at different latencies.}  
\label{unimodal_benchmark}
\end{table}

We conducted tests on the uni-modality based recognition at latencies of 0.2s, 0.5s, and 1s. The results, as shown in Table \ref{unimodal_benchmark}, reveal that the recognition accuracy is correlated with latency. As the latency increases, the classification performance for both modalities improves. This enhancement is attributed to the increased spatio-temporal information encapsulated within a longer period. At shorter latencies, the event modality exhibits markedly superior recognition accuracy compared to the frame. This superior performance is due to the higher temporal resolution of the event modality, which is adept at capturing rich motion information of the target even within brief time intervals (see Figure \ref{motion_sample}). This advantage becomes increasingly prominent as the latency decreases, showcasing the event modality’s substantial benefits in scenarios demanding rapid information processing.

\begin{table*}[htb]  
    \centering  
    \begin{tabular}{lllcccccccc}  
        \toprule  
        \multirow{2}{*}{Latency (s)} & \multirow{2}{*}{\( dt \) (ms)} & \multirow{2}{*}{\( T \)} & \multicolumn{1}{c}{\#EF} & \multicolumn{1}{c}{\#MF} & \multicolumn{2}{c}{\#LF} & \multirow{2}{*}{\#AF} & \multirow{2}{*}{Our CMA}\\  
        \cmidrule(lr){4-4} \cmidrule(lr){5-5} \cmidrule(lr){6-7}
        & & & Concat & Concat & OR & Average & & \\
        \midrule  
        \multirow{3}{*}{0.2} & 10 & 20  & 76.54 & 77.43 & 72.92 & 75.83 & 79.67 & \textbf{83.67}\\  
                      & 20 & 10  & 75.51 & 78.37 & 73.15 & 73.52 & 80.23 & \textbf{84.14}\\  
                      & 50 & 4   & 77.24 & 77.50 & 72.23 & 74.66 & 80.85 & \textbf{82.76}\\  
        \midrule  
        \multirow{3}{*}{0.5} & 10 & 50  & 77.29 & 78.79 & 76.32 & 75.91 & 80.84 & \textbf{83.17}\\  
                      & 20 & 25  & 77.10 & 78.87 & 78.42 & 77.65 & 82.57 & \textbf{86.97}\\  
                      & 50 & 10  & 78.62 & 80.67 & 74.22 & 77.43 & 81.45 & \textbf{85.46}\\  
        \midrule  
        \multirow{3}{*}{1}   & 20 & 50  & 79.41 & 79.79 & 78.10 & 80.41 & 82.44 & \textbf{85.71}\\  
                      & 50 & 20  & 80.37 & 80.18 & 80.37 & 78.19 & 83.10 & \textbf{87.23}\\  
                      & 100& 10  & 81.41 & 80.16 & 79.86 & 80.11 & 83.21 & \textbf{86.39}\\  
        \bottomrule  
    \end{tabular}  
\caption{Comparison of fusion accuracies at different latencies. \#EF denotes Early Fusion, \#MF denotes Middle Fusion, \#LF denotes Late Fusion, and \#AF represents the attention-based fusion method proposed by \citep{liu2022event}.} 
\label{dualmodal_comparision}  
\end{table*}

As noted above, the scarcity of dual-modality datasets has limited the development of SNN-based fusion methods. The work by \citep{liu2022event}, which integrates event and neuromorphic auditory modalities, has shown impressive performance in digit recognition. Given that their fusion module's input and output formats align with ours, we have included their algorithm in our comparative analysis. Additionally, we compared our method against other fusion methods, categorized by the timing of modality integration: early, middle, and late fusion. \textbf{(i) Early Fusion (EF)} is performed at the initial stages of the model, typically by concatenating the input vectors of each modality \citep{boulahia2021early, snoek2005early}. In our work, the implementation involves simply concatenating the input features \(X_e \in \mathbb{R}^{T \times 2 \times H \times W}\) and \(X_f \in \mathbb{R}^{T \times 3 \times H_0 \times W_0}\) to form \(X=[X_e; X_f] \in \mathbb{R}^{T \times 5 \times H_0 \times W_0}\). \textbf{(ii) Middle Fusion (MF)} allows the model to independently process the information of each modality to a certain level of abstraction before fusion, which can more effectively utilize the unique attributes of each modality \citep{li2018densefuse, nagrani2021attention}. Here, we concatenate the outputs of the SpikingConv Encoder for each modality, \(s^n=[s_e^n; s_f^n] \in \mathbb{R}^{T \times 2C \times H_n \times W_n}\), and subsequently input this combined data into the SpikingFC Decoder. \textbf{(iii) Late Fusion (LF)} occurs at the decision-making layer \citep{gadzicki2020early}. We employed two fusion approaches: performing an element-wise OR operation on the output spikes from each modality, and averaging the predicted probabilities.

\begin{figure*}[htb]
    \centering
    \includegraphics[scale=0.6]{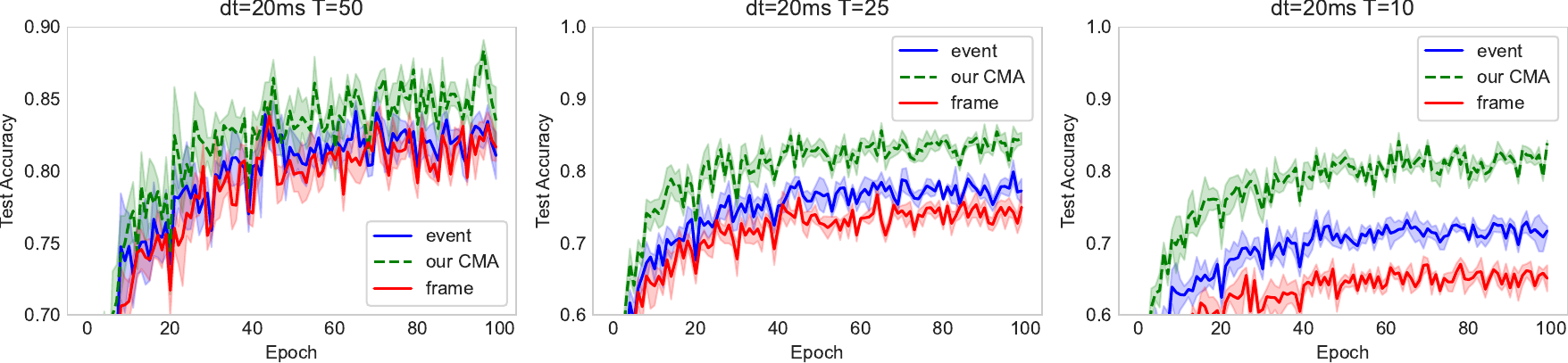}
    \caption{Test accuracy curves under various latency configurations. Here, \(dt\) is fixed at 20\(ms\). Each graph incorporates three curves, respectively representing frame based and event based uni-modality recognition, along with our CMA fusion.}
    \label{acc_curve}
\end{figure*}

As illustrated in Figure \ref{acc_curve}, across all latency parameters, the incorporation of the CMA module consistently enhances recognition accuracy compared to uni-modality recognition, with this improvement being particularly pronounced at lower latency settings. This is attributable to the frame modality's inability to provide sufficient dynamic information at such settings. In contrast, the CMA module effectively bridges the dynamic features from the event modality with the static attributes captured in the frame modality, thereby enriching the model's analytical depth and precision. The results of the dual-modality comparison experiments are presented in Table \ref{dualmodal_comparision}. It is evident that the attention-based fusion (AF) method, as proposed in \citep{liu2022event}, demonstrates superior performance over other simpler fusion approaches. This method simply computes attention scores for the spatio-temporal features of each modality, without adequately adapting to the unique characteristics inherent to each type of data. Our approach, on the other hand, employs Temporal-wise Attention (TA) and Spatial-wise Attention (SA) specifically for the event and frame modalities, respectively, thus more effectively leveraging the strengths of each modality, and consequently achieving optimal performance.

\subsubsection{Scenario Study}
\begin{table}[htb]
    \centering
    \begin{tabular}{llccc}
        \toprule
        \multicolumn{2}{c}{\multirow{2}{*}{Controlled Scenario}} & \multicolumn{2}{c}{Modality} & \multirow{2}{*}{Our CMA} \\
         \cmidrule(lr){3-4}
        & & Event  & Frame & \\
        \midrule
        \multirow{3}{*}{Light} & dimLED  & 76.29 & 66.75 & \textbf{81.56}\\
                               & brightLED  & 79.29 & 77.53 & \textbf{83.41}\\
                               & naturalLight   & 77.05 & 78.64 & \textbf{82.07}\\
        \midrule
        \multirow{2}{*}{Position} & back & 74.28 & 75.02 & \textbf{83.77}\\
                                  & front  & 82.59 & 80.51 & \textbf{84.49}\\
        \bottomrule
    \end{tabular}
    \caption{Recognition accuracy of event and frame streams under various lighting conditions and positions. \( dt \) and \( T \) are set to 50\(ms\) and 20, respectively.}
    \label{scenario_study}
\end{table}

Our dataset encompasses three distinct lighting conditions and two different positions. To analyze the unique characteristics of each modality and to verify the effectiveness of our fusion method, we extracted subsets from the full dataset for training and evaluation. The experimental results, as presented in Table \ref{scenario_study}, highlight several key findings: 
\textbf{(i) Superior Performance of Event Modality in Low-Light Conditions.} In low-light scenarios, the baseline model demonstrates a substantial decrease in recognition accuracy for the frame modality, while the event modality maintains a relatively high level of accuracy. This can be attributed to the unique generation mechanism of events. Unlike traditional RGB cameras that capture static light intensity information, event cameras only record changes in light intensity. Consequently, they have a higher dynamic range and are capable of accurately capturing and representing dynamic changes within scenes even under low-light conditions, thereby ensuring higher recognition accuracy. \textbf{(ii) The Influence of Position on Recognition Accuracy.} Datasets labeled as 'front' show enhanced recognition accuracy compared to those marked as 'back'. This difference is likely due to the superior signal-to-noise ratio found in the 'front' datasets. \textbf{(iii) Enhanced Model Robustness Through CMA Integration.} The incorporation of the CMA module has significantly improved accuracy across all scenarios, particularly notable in low-light conditions where the frame modality initially underperformed. Furthermore, the integration of the CMA module has led to a marked reduction in the discrepancy in accuracy across different conditions, indicating that the model has become less sensitive to varying scenarios and has achieved greater robustness.

\subsubsection{Ablation Study}
\begin{table}[htb]
    \centering
    \begin{tabular}{ccc}
        \toprule
            \multicolumn{2}{c}{Modality} & \multirow{2}*{Acc}\\
        \cmidrule{1-2}
        Event & Frame\\
        \midrule
        TA     & SA & \textbf{87.23}\\
        TA     & TA & 86.17\\
        SA     & SA & 85.71\\
        SA     & TA & 85.42\\
        \bottomrule
    \end{tabular}
\caption{Recognition accuracy comparison with different attention score generation strategies for each modality (\(dt = 50ms\), \(T = 20\))}
\label{module_ablation}
\end{table}

\noindent
\textbf{Impact of attention score generation strategy.} The CMA module functions by independently extracting temporal attention scores using temporal-wise attention (TA) and spatial attention scores using spatial-wise attention (SA), followed by their cross-fusion to produce enhanced features. We evaluated the effectiveness of this attention generation strategy. As shown in Table \ref{module_ablation}, the recognition accuracy was compared when applying different attention generation strategies to each modality. The results indicate that our method surpasses others in terms of accuracy, highlighting that such a combination more effectively utilizes the distinct characteristics inherent to each modality.

\begin{figure}[htb]
    \centering
    \includegraphics[scale=0.7]{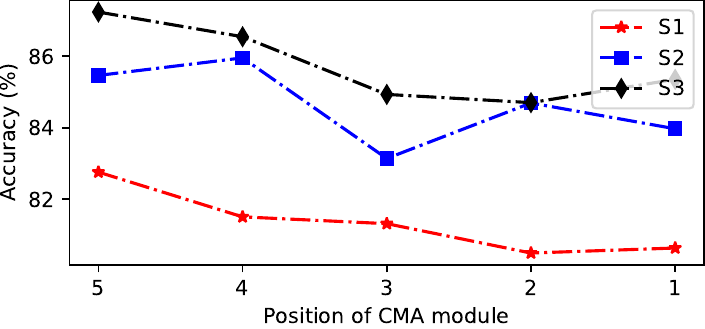}
    \caption{The relationship between different placement positions of the CMA module and accuracy. \(dt\) is set to 50ms. S1: \(T=4\), S2: \(T=10\), S3: \(T=20\). }
    \label{ablation}
\end{figure}

\noindent
\textbf{Impact of CMA module placement.} In our previous experiments, the CMA module was placed following the last convolutional block, setting \(n\) equal to \(N_{conv}=5\). However, the placement of the CMA module within the network can influence its performance. As evident from Figure \ref{ablation}, placing the CMA module in deeper layers slightly outperforms its placement in shallower layers. This finding aligns with observations in \citep{yao2021temporal, hu2018SE}, where attention modules applied to deep features also exhibited enhanced performance. Our experimental results show the effectiveness of fusion module placement.

\noindent
\textbf{Effectiveness of CMA module on different spiking neurons.} All experiments in this paper use the Leaky Integrate-and-Fire (LIF) neuron model. In this section, we discuss the performance of the CMA model on different neurons. We have selected the Integrate and Fire (IF), Parametric LIF (PLIF) \citep{fang2021incorporating}, and Leaky Integrate and Analog Fire (LIAF) \citep{wu2021liaf} neurons for evaluation. The IF neuron is a simplification of LIF, omitting the "leaky" process, meaning the time constant \(\tau\) is taken to be infinite. PLIF operates under the same computational paradigm as LIF, but with a trainable \(\tau\). LIAF updates the membrane potential in the same way as LIF, but uses an analog activation for output. Following \citep{wu2021liaf}, the scaled exponential linear unit (SELU) is applied as the activation function here. Table \ref{neuron_comparison} shows the recognition accuracies of these neurons using our CMA fusion and simple middle feature concat fusion respectively. It can be observed that without CMA, PLIF and LIAF achieve higher accuracies than LIF due to their more complex spatio-temporal representation. The CMA method enhances performance across all neuron types, but the improvement is smaller for LIAF. This may be due to the LIAF model using an analog activation for information transmission, while our CMA module is designed for spike-based outputs.

\begin{table*}[htb]
    \centering  
    \begin{tabular}{llcccccccc}  
        \toprule  
        \multirow{2}{*}{\( dt \) (ms)} & \multirow{2}{*}{\( T \)} & \multicolumn{8}{c}{Model Performance (With/Without CMA)} \\
        \cmidrule{3-10}
         & & \multicolumn{2}{c}{LIF} & \multicolumn{2}{c}{IF} & \multicolumn{2}{c}{PLIF} & \multicolumn{2}{c}{LIAF} \\
        \midrule
        \multirow{3}{*}{20} & 10  & \textbf{84.14} & 78.37 & \textbf{82.78} & 77.21 & \textbf{85.33} & 79.70 & \textbf{83.27} & 80.05 \\  
        
          & 25  & \textbf{86.97} & 78.87 & \textbf{84.01} & 77.53 & \textbf{86.74} & 79.63 & \textbf{84.10} & 81.73 \\  

          & 50  & \textbf{85.71} & 79.79 & \textbf{85.12} & 78.40 & \textbf{86.97} & 80.91 & \textbf{84.54} & 81.91 \\  

        \bottomrule  
    \end{tabular}  
\caption{Recognition accuracies of different spiking neuron models with and without CMA fusion.}
\label{neuron_comparison}
\end{table*}

\section{Conclusion}
In this work, we firstly verified the inadequacy of current neuromorphic datasets in terms of temporal correlation through Spike Timing Confusion and Temporal Information Elimination experiments. Addressing this, we designed and recorded the DVS-SLR dataset which exhibits high temporal correlation and effectively exploits the spatio-temporal processing capabilities of SNNs. Additionally, our neuromorphic dataset includes corresponding frame data, making it a valuable resource for developing SNN-based fusion techniques.

Secondly, we propose a Cross-Modality Attention (CMA)-based fusion approach. The CMA module efficiently utilizes the unique advantages of each modality, which enables SNNs to learn both temporal and spatial attention scores from the spatio-temporal features of event and frame modalities. Experimental results demonstrate that our method not only improves recognition accuracy but also ensures robustness across diverse scenarios.


\printcredits

\bibliographystyle{elsarticle-num-names}

\bibliography{cas-refs}

\end{document}